\documentclass[nohyperref]{article}

\usepackage{microtype}
\usepackage{graphicx}
\usepackage{booktabs} 

\usepackage[pagebackref]{hyperref}
\usepackage{soul}

\renewcommand*{\backref}[1]{}  
\renewcommand*{\backrefalt}[4]{
  \ifcase #1 
     No cited.
  \or
     (Cited on page #2.)
  \else
     (Cited on pages #2.)
  \fi}

\usepackage[accepted]{icml2022}

\usepackage{amsmath}
\usepackage{amssymb}
\usepackage{mathtools}
\usepackage{amsthm}

\usepackage{subcaption}
\usepackage{multirow}
\usepackage{wrapfig}
\usepackage{titlesec}
\mathtoolsset{showonlyrefs=true}
\usepackage[utf8]{inputenc} 

\usepackage{mathtools}
\DeclareMathOperator*{\argmax}{arg\,max}
\DeclareMathOperator*{\argmin}{arg\,min}
\mathtoolsset{showonlyrefs=true}

\def\tT{{\text{T}}}

\def\b0{{\boldsymbol{0}}}
 
\newcommand{\mE}{{\mathbb E}}
\newcommand{\mR}{{\mathbb R}}
\newcommand{\mA}{{\mathcal{A}}}
\newcommand{\mB}{{\mathcal{B}}}
\newcommand{\mD}{{\mathcal{D}}}
\newcommand{\mG}{{\mathcal{G}}}
\newcommand{\mP}{{\mathcal{P}}}
\newcommand{\mV}{{\mathcal{V}}}
\newcommand{\mL}{{\mathcal{L}}}
\newcommand{\mW}{{\mathcal{W}}}
\newcommand{\mN}{{\mathcal{N}}}

\newcommand{\mH}{{\mathcal{H}}}
\newcommand{\mF}{{\mathcal{F}}}
\newcommand{\mX}{{\mathcal{X}}}
\newcommand{\fan}[1]{\textcolor{black}{#1}}

\theoremstyle{plain}
\newtheorem{theorem}{Theorem}[section]
\newtheorem{proposition}[theorem]{Proposition}
\newtheorem{lemma}[theorem]{Lemma}
\newtheorem{corollary}[theorem]{Corollary}
\theoremstyle{definition}

\theoremstyle{remark}
\newtheorem{remark}[theorem]{Remark}
\usepackage[textsize=tiny]{todonotes}

\icmltitlerunning{Variational Wasserstein gradient flow}

\begin{document}

\twocolumn[
    \icmltitle{Variational Wasserstein gradient flow}



    

    \begin{icmlauthorlist}
        \icmlauthor{Jiaojiao Fan}{gatech}
        \icmlauthor{Qinsheng Zhang}{gatech}
        \icmlauthor{Amirhossein Taghvaei}{uw}
        \icmlauthor{Yongxin Chen}{gatech}        
    \end{icmlauthorlist}
   \icmlaffiliation{uw}{University of Washington, Seattle}
    \icmlaffiliation{gatech}{Georgia Institute of Technology}


    
    \icmlcorrespondingauthor{Jiaojiao Fan}{jiaojiaofan@gatech.edu}

    \icmlkeywords{Optimization, ICML}

    \vskip 0.3in
]



\printAffiliationsAndNotice{\icmlEqualContribution} 

\begin{abstract}
Wasserstein gradient flow has emerged as a promising approach to solve optimization problems over the space of probability distributions. A recent trend is to use the well-known JKO scheme in combination with input convex neural networks to numerically implement the proximal step. The most challenging step, in this setup, is to evaluate functions involving density explicitly, such as entropy, in terms of samples. This paper builds on the recent works with a slight but crucial difference: we propose to utilize a variational formulation of the objective function formulated as maximization over a parametric class of functions. Theoretically, the proposed variational formulation allows the construction of gradient flows directly for empirical distributions with a well-defined and meaningful objective function. Computationally, this approach replaces the computationally expensive step in existing methods, to handle objective functions involving density, with inner loop updates that only require a small batch of samples and scale well with the dimension. The performance and scalability of the proposed method are illustrated with the aid of several numerical experiments involving high-dimensional synthetic and real datasets.
\end{abstract}

\section{Introduction}

The Wasserstein gradient flow models the gradient dynamics on the space of probability densities with respect to the Wasserstein metric. It was first discovered by Jordan, Kinderlehrer, and Otto (JKO) in their seminal work \citep{JorKinOtt98}. They pointed out that the Fokker-Planck equation is in fact the Wasserstein gradient flow of the free energy,  bringing tremendous physical insights to this type of partial differential equations (PDEs). \fan{Since then, the Wasserstein gradient flow has played an important role in optimal transport~\citep{San17,CarDuvPeySch17}, PDEs~\citep{Ott01}, physics~\citep{CarCraWanWei21,AdaDirPelZim11}, machine learning~\citep{bunne2021jkonet,lin2021wasserstein,AlvSchMro21,FroPog20}, sampling~\citep{bernton2018langevin,cheng2018convergence,wibisono2018sampling} and many other areas~\citep{AmbGigSav08}.}
Despite the abundant theoretical results on the Wasserstein gradient flow established over the past decades \citep{AmbGigSav08,San17}, the computation of it remains a challenge. Most existing methods are either based on a finite difference method applied to  the underlying PDEs or based on a finite dimensional optimization; both require discretization of the underlying space \citep{Pey15,benamou2016discretization,CarDuvPeySch17,LiLuWan20,CarCraWanWei21}.
The computational complexity of these methods scales exponentially with the problem dimension, making them unsuitable for the cases with probability densities over high dimensional space.

{This shortcoming motivated recent line of interesting works to develop scalable algorithms utilizing neural networks~\citep{MokKorLiBur21,AlvSchMro21,YanZhaCheWan20,bunne2021jkonet,bonet2021sliced}. A central theme, in most of these works, is the application of the JKO scheme in combination with input convex neural networks (ICNN)~\citep{AmoXuKol17}. The JKO scheme, which is essentially a backward Euler method, is used to discretize the continuous flow in time. At each time-step, one needs to find a probability distribution that minimizes a weighted sum of squared Wasserstein distance, with respect to the distribution at the previous time-step, and the objective function. 
The probability distribution is then parametrized as push-forward of the optimal transport map from the previous probability distribution. The optimal transport map is represented with gradient of an ICNN utilizing the knowledge that optimal transport maps are gradient of convex functions when the transportation cost is quadratic. The problem is finally cast as stochastic optimization problem which only requires samples from the distribution.  

Our paper builds on these recent works but with a crucial difference. We propose to use a variational form of the objective function, leveraging $f$-divergences, which has been employed in multiple machine learning applications, such as generative models \citep{nowozin2016f}, and Bayesian inference \citep{wan2020f}. The variational problem is formulated as maximization over a parametrized class of functions. The variational form allows the evaluation of the objective in terms of samples,  without the need for density
estimation or approximating the logarithm of the determinant of the Hessian of ICNNs which appears in~\citep{MokKorLiBur21,AlvSchMro21}. Moreover, the variational form, even when restricted to a finite-dimensional class of functions, admits nice geometrical properties of its own leading to a  meaningful objective function to minimize.}

At the end of the algorithm, a sequence of transport maps connecting the initial distribution with the terminal distribution along the gradient flow dynamics are obtained. One can then sample from the distributions along the flow by sampling from the initial distribution (often Gaussian) and then propagating these samples through the sequence of transport maps. When the transport map is modeled by the gradient of an input convex neural network, one can 
{also} evaluate the densities at every point.

Our contributions are summarized as follows.\\
i) We develop a 
numerical algorithm to implement the Wasserstein gradient flow that is based on a variational representation of the objective functions. The algorithm does not require spatial discretization, density estimation, or approximating logarithm of determinant of Hessians. 
\\
ii) We numerically demonstrate the performance of  our algorithm on several representative problems including sampling from high-dimensional Gaussian mixtures, porous medium equation, and learning generative models on MNIST and CIFAR10 datasets. We illustrate the computational advantage of our proposed method in comparison with~\citep{MokKorLiBur21,AlvSchMro21}, in terms of computational time and scalibity with the problem dimension.\\
iii) We establish some preliminary theoretical results regarding the proposed variational objective function. In particular, we provide conditions under which the variational objective satisfies a moment matching property and an embedding inequality with respect to a certain integral probability metric (see Proposition~\ref{lem:DfH}).

    {\bf Related works:} Most existing methods to compute Wasserstein gradient flow are finite difference based \citep{Pey15,benamou2016discretization,CarDuvPeySch17,LiLuWan20,CarCraWanWei21}. These methods require spatial discretization and are thus not scalable to high dimensional settings. 
    There is a line of research that uses particle-based method to estimate the Wasserstein gradient flow \citep{CarCraPat19,FroPog20}. In these algorithms, the current density value is often estimated using kernel method whose complexity scales at least quadratically with the number of particles. More recently, several interesting neural network based methods \citep{MokKorLiBur21,AlvSchMro21,YanZhaCheWan20,bunne2021jkonet,bonet2021sliced,hwang2021deep} were proposed for Wasserstein gradient flow.
\citet{MokKorLiBur21} focuses on the special case with Kullback-Leibler divergence as objective function.
\citet{AlvSchMro21} uses a density estimation method to evaluate the objective function by back-propagating to the initial distribution, which could become a computational burden when the number of time discretization is large.
\citet{YanZhaCheWan20} is based on a forward Euler time discretization of the Wasserstein gradient flow and is more sensitive to time stepsize.
    {\citet{bunne2021jkonet} utilizes JKO scheme to approximate a population dynamics given an observed trajectory, which finds application in computational biology. \citet{bonet2021sliced} replaces Wasserstein distance in JKO by sliced alternative but its connection to the original Wasserstein gradient flow remains unclear.}


\section{Background}

\subsection{Optimization problem }\label{sec:problem}
 We are interested in developing algorithms
for 
\begin{equation}\label{eq:OPT}
    \min_{P \in \mathcal P_{ac}(\mathbb{R}^n)}  ~\mathcal{F} (P),
\end{equation}
where $\mathcal P_{ac}(\mathbb{R}^n)$ is
the space of probability distributions that
admit density $dP/dx$ with respect to Lebesgue measure.
The objective function $\mathcal{F} (P)$ takes different form depending on the application.
Three important examples are:  

\textbf{Example I:} Kullback-Leibler divergence with respect to a given target distribution $Q$,
\begin{equation}\label{eq:kl}
    \mD(P||Q) := \int \log \left(\frac{dP}{dQ} \right)dP
\end{equation}
plays an important role in 
the sampling problem. 

\textbf{Example II:} 
Generalized entropy
\begin{equation}
    \mG(P):= \frac{1}{m-1}\int P^m(x) dx,\quad m>1 
\end{equation}
is important for modeling the porous medium.

\textbf{Example III:} The (twice) Jensen-Shannon divergence
\begin{align}
     {\rm JSD}(P\|Q):= \mD \left(P \left\Vert \frac{P+Q}{2} \right. \right) +  \mD \left(Q \left \Vert \frac{P+Q}{2} \right. \right)
\end{align}
is
important in learning generative models. 

\subsection{Wasserstein gradient flow}
Given a function $\mF(P)$ over the space of probability densities, the Wasserstein gradient flow describes the dynamics of the probability density when it follows the steepest descent direction of the function $\mF(P)$ with respect to the Riemannian metric induced by the $2$-Wasserstein distance $W_2$~\cite{AmbGigSav08}. The Wasserstein gradient flow can be explicitly represented by the PDE
\begin{align}
    \frac{\partial P}{\partial t} = \nabla \cdot \left(P \nabla \frac{\delta \mF}{\delta P}\right),
\end{align}
where $\delta \mF/\delta P$ represents the first-variation of
of $\mF$ with respect to the standard $L_2$ metric~\citep[Ch. 8]{Vil03}.

Wasserstein gradient flow corresponds to various important PDEs  depending on the choice of objective functions $\mF(P)$. For instance, when $\mF(P)$ is the free energy, i.e. 
\begin{equation}
  \mF (P) = \int_{\mR^n} P(x) \log P(x) dx + \int_{\mR^n} V(x) P(x) dx,
\end{equation}
the gradient flow is the Fokker-Planck equation~\citep{JorKinOtt98}. 
\begin{equation}
  \frac{\partial P}{\partial t} = \nabla \cdot(P\nabla V) + \Delta P.
\end{equation}
When $\mF(P)$ is the generalized entropy
$ \mF(P) = \frac{1}{m-1}\int_{\mR^n} P^m(x) dx$
for some positive number $m>1$,
the gradient flow is the porous medium equation~\citep{Ott01,Vaz07}
$  \frac{\partial P}{\partial t} = \Delta P^m$. 


\subsection{JKO scheme and reparametrization}\label{sec:jko}
To numerically realize the Wasserstein gradient flow, a discretization over time is needed. One such discretization is the famous JKO scheme \citep{JorKinOtt98}
\begin{equation}\label{eq:proximal}
    P_{k+1}=\argmin_{P \in 
    {\mP_{ac}}(\mathbb{R}^n)} \frac{1}{2a} W_2^2 \left( P, P_k\right)+\mathcal{F}(P).
\end{equation}
This is essentially a backward Euler discretization or a proximal point method with respect to the Wasserstein metric. 
The solution to \eqref{eq:proximal} converges to the continuous-time Wasserstein gradient flow when the step size $a \rightarrow 0$.

Recall the definition of the Wasserstein-2 distance
\begin{equation}\label{eq:OTform} 
   W_2^2(P,Q) =\min_{T: T\sharp P = Q} \int_{\mR^n} \|x-T(x)\|_2^2 dP(x),
\end{equation}
where the minimization is over all the feasible transport maps that transport mass from distribution $P$ to distribution $Q$.
{Hence,} \eqref{eq:proximal} can be recast as an optimization in terms of the transport maps $T: \mR^n \rightarrow \mR^n$ from $P_k$ to $P$. By defining $P=T\sharp P_k$, the optimal $T$ is the optimal transport map from $P_k$ to $T\sharp P_k$ and thus the gradient of a convex function $\varphi$ 
{by Brenier's Theorem \citep{brenier1991polar}.} 
    \citet{bunne2021jkonet,MokKorLiBur21,AlvSchMro21} propose to parameterize $T$ as the gradient of Input convex neural network (ICNN) \citep{AmoXuKol17} and express \eqref{eq:proximal} as
        \begin{align}\label{eq:JKO-phi}
             & P_{k+1} = \nabla \varphi_k\sharp P_k,                                                                                                              \\
             & \varphi_k=\argmin_{\varphi \in \text{CVX}} \frac{1}{2a} \int_{\mR^n} \|x\!-\!\nabla \varphi(x)\|_2^2 dP_{k}(x) \!+\!\mathcal{F}(\nabla \varphi\sharp P_k),
        \end{align}
        where $\text{CVX}$ stands for the space of convex functions.
        In our method, we extend this idea and  propose to reparametrize $T$ alternatively by a residual neural network.
        With this reparametrization, the JKO step \eqref{eq:proximal} becomes
        \begin{align}\label{eq:JKO-T}
             & P_{k+1} = T_k\sharp P_k,                                                                      \\
             & T_k=\argmin_{T} \frac{1}{2a} \int_{\mR^n} \|x\!-\!T(x)\|_2^2 dP_{k}(x) \!+\!\mathcal{F}(T\sharp P_k).
        \end{align}
        We use the preceding two schemes \eqref{eq:JKO-phi} and \eqref{eq:JKO-T} in our numerical method depending on the application.

\section{Methods and algorithms} \label{sec:method}
We discuss how to implement JKO scheme with our approach and its computational complexity in this section.

\subsection{ $\mF(P)$
reformulation with variational formula} \label{sec:var_form}
The main challenge in implementing the JKO scheme is to evaluate the functional $\mF(P)$ in terms of samples from $P$.
We achieve this goal by using a variational formulation of $\mF$.
In order to do so, we use the notion of $f$-divergence
between the two distributions $P$ and $Q$:
\begin{align}
    \label{eq:f-divergence}
    D_f(P\| Q)= \mE_Q \left[ f \left( \frac{dP}{dQ}\right)\right]
\end{align}
where $P$ admits density with respect to $Q$ (denoted as $P \ll Q$) and $f:  [0,+\infty) \to \mathbb{R}$ is a convex and lower semi-continuous function. Without loss of generality, we assume $f(1)=0$ so that $D_f$ attains its minimum at $P=Q$.
\begin{proposition}\label{lem:var} \citep{nguyen2010estimating}
$\forall P,Q \in \mP(\mR^n)$ such that $P \ll Q$ 
and differentiable $f$:
\begin{align}
    \label{eq:f-divergence-dual}
    D_f(P\| Q)= \sup_{h \in \mathcal{C}}   \mE_P [h(X)]-\mE_Q [f^*(h(Y))].
\end{align}
where {$f^*(y)=\sup_{x \in \mR} [xy -f(x)]$ is the convex conjugate of $f$ and $\mathcal C$ is all measurable functions $h: \mR^n \to \mR$.
} The supremum is {attained at $h =f'(dP/dQ)$.}
\end{proposition}
The variational form has the 
{distinguishing} feature that it does not involve the density of $P$ and $Q$ explicitly and can be approximated in terms of samples from $P$ and $Q$.
In general, our scheme can be applied to any $f$-divergence, but we focus on the functionals in Section \ref{sec:problem}. 

With the help of the  $f$-divergence variational formula, when $\mF(P)=\mD (P\|Q)$, $\mG(P)$ or JSD$(P\|Q)$, 
the JKO scheme~\eqref{eq:JKO-T} can be equivalently expressed as
\begin{align}\label{eq:saddle}
     & P_{k+1}=T_k\sharp P_k,                                                        \\
     & T_k = \argmin_T \left\{ \frac{1}{2a} \mathbb{E}_{P_k}[\|X-T(X)\|^2] +  \sup_h
    \mV(T,h)
    \right\}.
\end{align}
where $\mV(T,h)=\mE_{X \sim P_k}[\mathcal{A}_h(T(X))] - \mE_{Z \sim \Gamma}[\mathcal{B}_h (Z)]$, $\Gamma$ is a user designed distribution which is easy to sample from, and $\mA$ and $\mB$ are functionals whose form depends on $\mF$. The specializations of $\mA$ and $\mB$ appear in Table~\ref{table}. 

The following lemma implies that if $ \mF(P) $ can be written as $D_f(P \| Q)$, then $\mF(P)$ monotonically decreases along its Wasserstein gradient flow, which makes it reasonable to solve \eqref{eq:OPT} by using JKO scheme. It also justifies that the gradient flow finally converges to $Q$.
\begin{lemma} \citet[Lemma 2.2]{gao2019deep} \\
$\frac{d}{dt} \mF(P_t) = 
{- \mE_{P_t}(\|\nabla f'(P_t/Q) \|^2 )} $.
\end{lemma}


\subsubsection{KL divergence}\label{sec:KL}
The KL divergence is a special instance of the $f$-divergence with $f(x) = x \log x$. 
\fan{Using $f(x) = x\log x$ in \eqref{eq:f-divergence-dual} yields the following expression for KL divergence as a corollary of Proposition \ref{lem:var}. The proof appears in Section \ref{sec:detail_var} }
\begin{corollary}\label{cor:kl}
The variational form for $\mD(P\|Q)$ reads
    \begin{equation}\label{eq:var_kl}
        \mD(P\|Q)\!=\!1+\sup_h \mathbb{E}_{P}\!\left[\log\frac{h(X) \mu (X)}{Q(X)} \right] - \mE_\mu \left[{h(Z)}\right],
    \end{equation}
    where $\mu$ is a user designed distribution which is easy to sample from.
    The optimal function $h$ is equal to $dP/d\mu$.
\end{corollary}
This variational formula
becomes practical when we have only access to un-normalized density of $Q$, which is the case for the sampling problem. 
In practice, we choose $\mu=\mu_k$ adaptively, where $\mu_k$ is the Gaussian with the same mean and covariance as $P_{k}$. 
We noticed that this choice improves the numerical stability of the the algorithm.

\subsubsection{{Generalized entropy}}\label{sec:porous-f}
The generalized entropy can be also represented as a $f$-divergence. In particular, with $f(x)=\frac{1}{m-1} (x^m-x)$ and $Q$ the uniform distribution on the superset of the support of density $P(x)$ with volume $\Omega$:
\begin{align}\label{eq:f_diverg_porous}
    D_{f}(P\| Q)&=\frac{\Omega^{m-1}}{m-1} \int P^m(x)dx -\frac{1}{m-1} \\&= \Omega^{m-1}\mathcal G(P) - \frac{1}{m-1} .
\end{align}
\fan{Plugging $f(x) = \frac{1}{m-1} (x^m-x) $ into \eqref{eq:f-divergence-dual}, we get the following expression of the generalized entropy as a corollary of Proposition \ref{lem:var}. The proof appears in Section \ref{sec:detail_var} }
\begin{corollary}\label{cor:porous}
    The variational formulation for $\mG(P)$ reads
    \begin{align}\label{eq:porous_hat_h}
        \!\!\!\mG(P)\!=\!\frac{\sup_{ {h}} \left( \mE_{P} \left[\frac{m {h}^{m-1}(X)}{m-1}    \right]-\mE_Q \left[ {h}^m(Z)  \right] \right)}{\Omega^{m-1}}.
    \end{align}
    The optimal function $h$ is equal to $dP/dQ$.
\end{corollary}
In practice, we choose $\Omega=\Omega_k$ 
which is the volume of a set that guarantees to contain the support of $T \sharp P_k$.
In view of the connection between generalized entropy and $f$-divergence, it is justified that the solution of Porous Media equation develops towards a uniform distribution. Especially, when $m=2$, \eqref{eq:f_diverg_porous} recovers
    the Pearson divergence between $P$ and the uniform distribution $Q$. 
\subsubsection{Jensen-Shannon divergence}\label{sec:js}
JSD$(P\|Q)$ corresponds to $f$-divergence with $f(x)=-(x+1)\log((1+x)/2) +x\log x$. \fan{Direct application of \eqref{eq:f-divergence-dual} concludes the following Corollary.}
\begin{corollary}\label{prop:js}
    The variational form for JSD$(P\|Q)$ is
    \begin{align}\label{eq:js}
   \!\!\!\!\! \log 4 + \sup_{ {h}} \mE_{P} \left[\log(1-h(X)) \right] + \mE_Q \left[\log {h}(Z)  \right].
    \end{align}
\end{corollary}
In particular, we apply JSD to the learn the image generative model, therefore we assume 
{samples from $Q$ are accessible.}
\begin{table}[]
    \caption{Variational formula for $\mF(P)$ }
    \begin{center}
        \begin{small}
            \begin{tabular}{cccc}
                \toprule
                $\mF(P)$     & $\mA_h $                                                             & $\mB_h $     & $\Gamma$                      \\
                \toprule
                $\mD(P\|Q)$     & $ \log \left(\frac{h \cdot \mu_k}{Q}\right) $       & $h$           & Gaussian dist. $\mu_k$ \\
                \midrule
                $\mG(P)$ & $\frac{m}{m-1}  \cdot \frac{h^{m-1}}{\Omega_k^{m-1}} 
                $                           & $ \frac{h^m}{\Omega_k^{m-1}} $                          & Uniform dist. $Q_k$                                 \\
                \midrule
                JSD$(P\|Q)$ & $\log(1-h)
                $                           & $ -\log h$                          & Empirical dist. $Q$                                 \\
                \bottomrule
            \end{tabular}
        \end{small}
    \end{center}
    \vskip -0.1in
    \label{table}
\end{table}
\begin{algorithm}[H]
    \caption{Primal-dual gradient flow}
    \label{algo:flow}
    \begin{algorithmic}
        \STATE{{\bfseries Input:} Objective functional $\mF(P)$, initial distribution $P_0$, JKO step size $ a$, number of JKO steps $K$.
        }
        \STATE{{\bfseries Initialization:} Parameterized $T_\theta$ and $h_\lambda$
        }
        \FOR{ $k=1,2,\ldots, K$}
        \STATE $T_\theta \leftarrow T_{k-1}$ if $k>1$  $\qquad$
        \FOR{ $j_1=1,2,\ldots, J_1$}
        \STATE Sample $X_1,\ldots, X_M \sim P_k$,  $~ Z_1,\ldots, Z_M \sim \Gamma$.
        \STATE Maximize $\frac{1}{M} \sum_{i=1}^M \left[ \mA(T_\theta (X_i), h_\lambda ) - \mB(h_\lambda (Z_i)) \right]$ over $\lambda$ for $J_2$ steps.
        \STATE Minimize $
            \frac{1}{M}\! \sum_{i=1}^M
            \!\!\left[ \frac{\|X_i - T_\theta (X_i)\|^2}{2a}
                \!+\!\mA(T_\theta(X_i), h_\lambda ) \right]$ over $\theta$ for $J_3$ steps.
        \ENDFOR
        \STATE $ T_k \leftarrow T_\theta$
        \ENDFOR
        \STATE{{\bfseries Output: $\{T_k\}_{k=1}^K$}}
    \end{algorithmic}
\end{algorithm}

\subsection{
    Parametrization of $T$ and $h$
}\label{sec:algo}
The two optimization variables $T$ and $h$ in our minimax formulation \eqref{eq:saddle} can be both parameterized by neural networks, denoted by $T_\theta$ and $h_\lambda$. With this neural network parametrization, we can then solve the problem by iteratively updating $T_\theta$ and $h_\lambda$. This primal-dual method to solve \eqref{eq:OPT} is depicted in Algorithm \ref{algo:flow}.

In this work, we implemented two different architectures for the map $T$. One way is to use a {residual} neural network to represent $T$ directly, and another way is to parametrize $T$ as the gradient of a ICNN $\varphi$. The latter has been widely used in optimal transport \citep{MakTagOhLee20,FanTagChe20,KorLiSolBur21}.
However, recently several works \citep{rout2021generative,KorEgiAsaBur19,fan2021monge,bonet2021sliced} find poor expressiveness of ICNN architecture and also propose to replace the gradient of ICNN by a neural network. In our experiments, we find that the first parameterization gives more regular results, which aligns with the result in \citet[Figure 4]{bonet2021sliced}. However, it would be very difficult to calculate the density of pushforward distribution. Therefore, with the first parametrization, our method becomes a particle-based method, i.e. we cannot query density directly.
As we discuss in Section \ref{sec:density}, when density evaluation is needed, we adopt the ICNN since we need to compute $T^{-1}$.


\subsection{Computational complexity}\label{sec:complexity}
Each update $k$ in Algorithm \ref{algo:flow} 
requires {$O(J_1 k M  H)$} operations, where $J_1$ is the number of iterations per each JKO step, $M$ is the batch size, and $H$ is the size of the network.
$k$ shows up in the bound because sampling $P_k$ requires us to pushforward $x_0 \sim P_0$ through $k-1$ maps.

In contrast,
\citet{MokKorLiBur21}  requires {$O \left(J_1  \left( (k+n) M  H  +n^3 \right) \right)$} operations, which has a cubic dependence \citep[Section 5]{MokKorLiBur21} on dimension $n$ because they need to query the $\log \det \nabla^2 \varphi$ in each iteration.
    {There exists fast approximation \citep{huang2020convex} of $\log \det \nabla^2 \varphi$ using Hutchinson trace estimator \citep{hutchinson1989stochastic}.
        \citet{AlvSchMro21} applies this technique, thus the cubic dependence on $n$ can be improved to quadratic dependence. Noneless, this is accompanied by an additional cost, which is the number of iterations to run conjugate gradient (CG) method. CG is guaranteed to converge  exactly in $n$ steps in this setting. If one wants to obtain $\log \det \nabla^2 \varphi$ precisely, the cost is still $O(n^3)$, which is the same as calculating $\log \det \nabla^2 \varphi$ directly. 
        If one uses an error $\epsilon$ stopping condition in CG, the complexity could be improved to $\sqrt{\kappa} \log (2 / \epsilon) n^2$ \cite{shewchuk1994introduction}, where $\kappa$ is the upper bound of condition number of $\nabla^2 \varphi$, 
        but this would sacrifice on the accuracy.}
Given the similar neural network size, our method has the advantage of independence on the dimension for the training time.

Other than training time, the complexity for evaluating the density has unavoidable dependence on $n$ due to the standard density evaluation process (see Section \ref{sec:density}).
\vspace{-0.2cm}
\section{Theoretical results}\label{sec:theory}
We introduce approximate $f$-divergence notation and analyze its properties in this section. 

\subsection{Approximate $f$-divergence}
Given the results in Proposition \ref{lem:var},
 now we consider a restriction of the optimization domain  $\mathcal C$ to a class of functions $\mathcal H$, e.g parametrized by neural networks, and define the new functional 
\begin{equation}\label{eq:DHf}
    D^{\mathcal H}_f(P\|Q)= \sup_{h \in \mathcal H}\left\{\int h d P - \int f^*(h)dQ\right\}.
\end{equation}
This functional forms a surrogate for the exact $f$-divergence. 
It is straightforward to see that the new function is always smaller than the exact $f$-divergence, i.e. $D^{\mathcal H}_f(P\|Q)\leq D_f(P\|Q)$ where the inequality is achieved when $f'(\frac{dP}{dQ})$ belongs to $\mathcal H$. In the following lemma, we establish some important theoretical properties of the approximate $f$-divergence $D^{\mathcal H}_f(P\|Q)$. In order to do so, we introduce the integral probability metric~\citep{sriperumbudur2012empirical,arora2017generalization}   
\begin{equation*}
     d_{\mathcal H}(P,Q) =   \sup_{h \in \mathcal H} \frac{1}{\|h\|_{2,Q}}\left\{ \int hdP - \int h dQ\right\},  
    \end{equation*}
where $\|h\|_{2,Q}^2 =\int h^2 dQ$.

\begin{proposition}\label{lem:DfH}
The approximate $f$-divergence $D^{\mathcal H}_f(
P\|Q)$ satisfies the following properties:\\ 
    1. (positivity) If $\mathcal H$ contains all constant functions, then 
    \begin{equation*}
        D_f^{\mathcal H}(P\|Q) \geq 0,\quad \forall P,Q.
    \end{equation*}
    2. (moment-matching) If for all $h \in \mathcal H$, $c + \lambda h \in \mathcal H$ for $c,\lambda \in \mathbb{R}$, then
    \begin{equation*}
        D_f^{\mathcal H}(P\|Q) = 0 ~~ \Leftrightarrow ~~ \int h dP = \int h dQ,~\forall h \in \mathcal H.
           \end{equation*}
    3. (embedding inequalities) Additionally, if $f$ is strongly convex with constant $\alpha$, and smooth with constant $L$, then,
    \begin{equation*}
\frac{\alpha}{2}d_{\mathcal H}(P,Q)^2  \leq  D_f^{\mathcal H }(P\|Q) \leq  \frac{L}{2}  d_{\mathcal H}(P,Q)^2. 
    \end{equation*}
\end{proposition}

The proposition has important implications. Part (1) establishes the condition under which the approximate $f$-divergence is always positive. Part (2) identifies necessary and sufficient conditions under which the approximate divergence is zero for two given probability distributions $P$ and $Q$. In particular, the divergence is zero iff the moments of $P$ and $Q$  are equal for all functions in the function class $\mathcal H$. Finally, part (3) provides lower-bound and upper-bound for the approximate $f$-divergence in terms of an integral probability metric defined on the function class $\mathcal H$, implying that the two measures are equivalent \fan{when $f$ is both strongly convex and smooth}. For example,  a sequence $D_f^{\mathcal H}(P_d \| Q_d) \to 0$ as $d \to \infty$ iff $d_{\mathcal H}(P_d,Q_d) \to 0$ as $d \to \infty$. 
Or if we are able to minimize the approximate $f$-divergence $D_f^{\mathcal H}(P\|Q)$ with optimization gap $\epsilon$, then the error in the moments of $P$ and $Q$ for functions in $\mathcal H$ is of order $O(\sqrt{\epsilon})$. These results inform us that the proposed objective function of minimizing $D_f^{\mathcal H}(P\|Q)$ is meaningful and has geometrical significance.  

\begin{remark} The assumption that $c+\lambda h\in \mathcal H$ for all $h\in \mathcal H$ and $c,\lambda \in \mathbb{R}$ holds for any neural network with linear activation function at the last layer. The assumption that $f$ is strongly convex and smooth may not hold for a typical $f$ such as $f(x)=x\log(x)$ over $(0,\infty)$. However, It holds when the domain is restricted, which is true when either the samples are bounded or $h$ is bounded for all $h\in \mathcal H$.
\end{remark}
\subsection{Computational boundness}\label{sec:error-analysis}
It is also possible to obtain lower-bound for $D_f^{\mathcal H}(P\|Q)$ in terms of the exact $f$-divergence $D_f(P\|Q)$ when the class $\mathcal H$ is rich enough. 
\begin{proposition}\label{lem:low_bound}
If $f$ is $\alpha$-strongly convex and the class of functions is able to approximate any function $h\in \mathcal C$ with $\tilde{h} \in \mathcal H$ such that $\|\tilde{h}-h\|_{2,Q}\leq \epsilon$, then 
\begin{equation*}
    D_f^{\mathcal H}(P\|Q) \geq  D_f(P\|Q) - \frac{\epsilon^2}{2\alpha},\quad \forall P,Q.
\end{equation*}
\end{proposition}
\fan{Proposition \ref{lem:low_bound} gives upper-bound on the error between variational $f$-divergence and the ground truth by the function class expressiveness, which can be verified for neural net function class. Assume $\mH$ is the class of neural nets with an arbitrary depth under mild assumption on the activation function. Following the proof of Theorem 1 in ~\citet{Korotin2022NeuralOT}, we can verify that for any $\epsilon>0$, compactly supported $Q$, and function $\|h \|_{2,Q} < \infty $, there exists a neural net $\tilde{h} \in \mH$ such that $\|\tilde{h}-h\|_{2,Q}\leq \epsilon$ (c.f. discussion in Section \ref{sec:low_bound_proof}). 
However, Proposition \ref{lem:DfH}-(3) and Proposition \ref{lem:low_bound} require $f$ to be strongly convex, which might be too strong for some $f$-divergences, such as KL divergence. }

Unlike the exact form of the $f$-divergence, the variational formulation is well-defined for empirical distributions when the function class $\mathcal H$ is restricted and admits a finite Rademacher complexity. 
\begin{proposition}\label{lem:rad}
Let $P^{(N)} = \frac{1}{N}\sum_{i=1}^N \delta_{X_i}$, $Q^{(M)}= \frac{1}{M}\sum_{i=1}^M \delta_{Y_i}$, where $\{X_i\}_{i=1}^N, \{Y_i\}_{i=1}^M$ are i.i.d samples from $P$ and $Q$ respectively. Then, it follows
that
\begin{align*}
    & \mathbb E[|D^{\mathcal H}_f(
P\|Q) - D^{\mathcal H}_f(
P^{(N)}\|Q^{(M)})|] \\
\leq & 2\mathcal R_N(\mathcal H,P) +2\mathcal R_M(f^* \circ \mathcal  H,Q) ,
\end{align*}
where the expectation is over the samples and  $\mathcal R_N(\mathcal H,P)$ denotes the Rademacher complexity of the function class $\mathcal H$ with respect to $P$ for sample
size $N$. 
\end{proposition}
Proposition \ref{lem:rad} quantifies the generalization error in terms of Rademacher complexity.
We leave the task of evaluating the Rademacher complexity for different function classes employed in this paper for future work. 

\subsection{Convergence to spherical Gaussian distribution}
We assert the efficacy of JKO with variational estimation through
a spherical Gaussian example. We consider  sampling from the target distribution $Q=\mN (\eta,I_n)$ by minimizing the functional $\mF(P) = \mD (P \|Q).$ We choose $P_0 = \mu = \mN(0, I_n)$, and parameterize $T$ to be linear functions. 
Assume we get $T_0,\ldots,T_{K-1}$ by solving the particle approximated JKO in \eqref{eq:kl_jko_sample}, and we can estimate $\mE_\mu [h(\cdot)]$ precisely for simplication. Denote
$P_{K}$ as the $K$-th JKO iteration $T_{K-1} \sharp (\ldots (T_{0} \sharp P_0) )$ and $P^*_{K}$ as the ground truth solution of JKO.
\begin{proposition}\label{prop:gauss}
Based on the assumptions in the paragraph above, let $P^{(N)}_K = \frac{1}{N}\sum_{i=1}^N \delta_{X_i}$, where $\{X_i\}_{i=1}^N$ are i.i.d samples from $P_{K}$. Then, it follows
that
\begin{align*}
& \mathbb E[|\mD^{\mathcal H}(
P_K^* \|Q) - \mD^{\mathcal H} (
P^{(N)}_K \|Q)|] \\
\leq &  \Delta_K \sqrt{\xi_{K,N} } + {\xi_{K,N}}/{2 } 
\end{align*}
where $\Delta_K = \frac{\|\eta\|}{ (1+a)^{ K} } ,$ $$\xi_{K,N} = \left(\frac{a}{1+a} \right)^2 \frac{n}{N} \sum_{j=1}^K \frac{1}{ (1+a)^{2(K-j)} } + \frac{n}{N} ,$$
 and $\mH \supseteq \{h: h(z) = \exp(\alpha^\top z + \gamma ), \alpha \in \mR^n, \gamma \in \mR \}  $.
\end{proposition}
This proposition quantifies the sample complexity and convergence rate of  JKO with our variational estimation for a spherical Gaussian example.
In the future, it would be useful to analyze the stability and convergence of the proposed min-max formulation for more general functional $\mF(P)$, both at the level of densities and at the level of samples/particles.

\section{Numerical examples} \label{sec:examples}
In this section, we present several numerical examples to illustrate our algorithm.
We mainly compare with the JKO-ICNN-d \citep{MokKorLiBur21}, JKO-ICNN-a \citep{AlvSchMro21}.
The difference between JKO-ICNN-d and JKO-ICNN-a is that the former computes the $\log \det (\nabla^2 \varphi)$ \textbf{d}irectly and the latter adopts fast \textbf{a}pproximation. 
We use the default hyper-parameters in the authors' implementation.  \fan{Our code is written in
PyTorch-lightning and is publicly available at \url{https://github.com/sbyebss/variational_wgf}.}

\subsection{Sampling from Gaussian Mixture Model} \label{sec:sampling}
\begin{figure}[h]
    \centering
    \begin{subfigure}{0.4\textwidth}
        \includegraphics[width=1\linewidth]{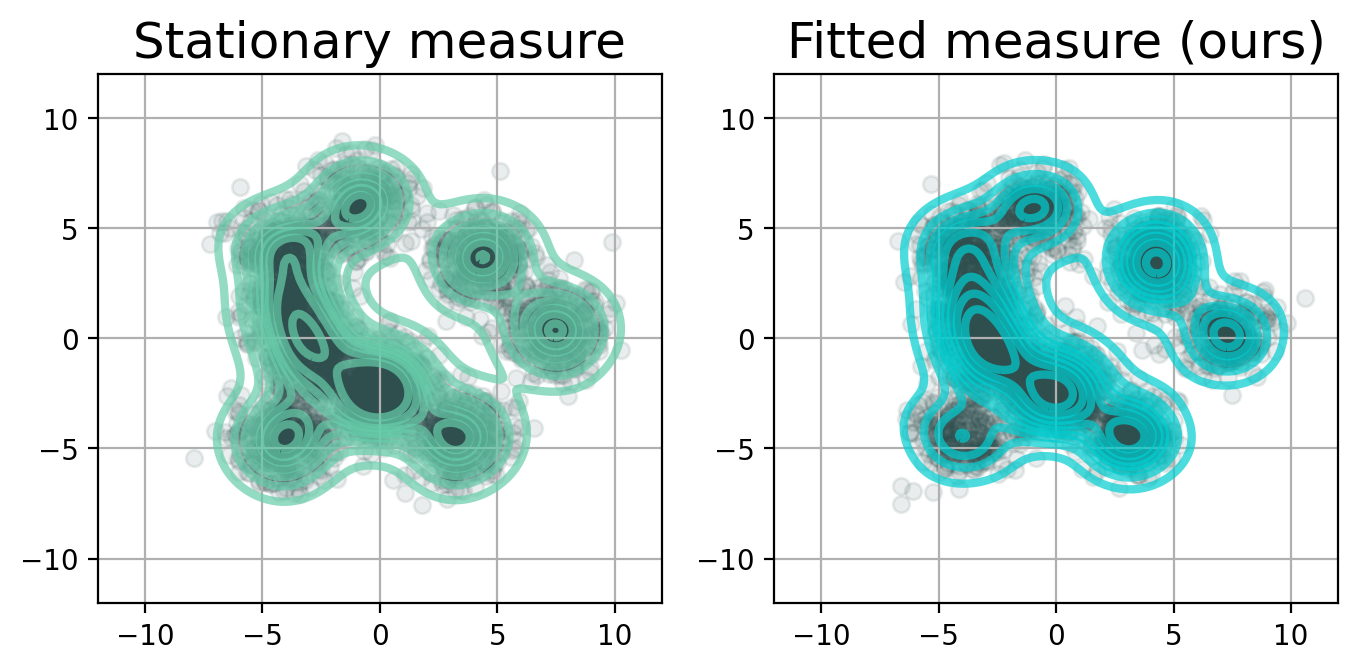}
        \caption{Dimension $n=64$}
    \end{subfigure}
    
    \begin{subfigure}{0.4\textwidth}
        \includegraphics[width=1\linewidth]{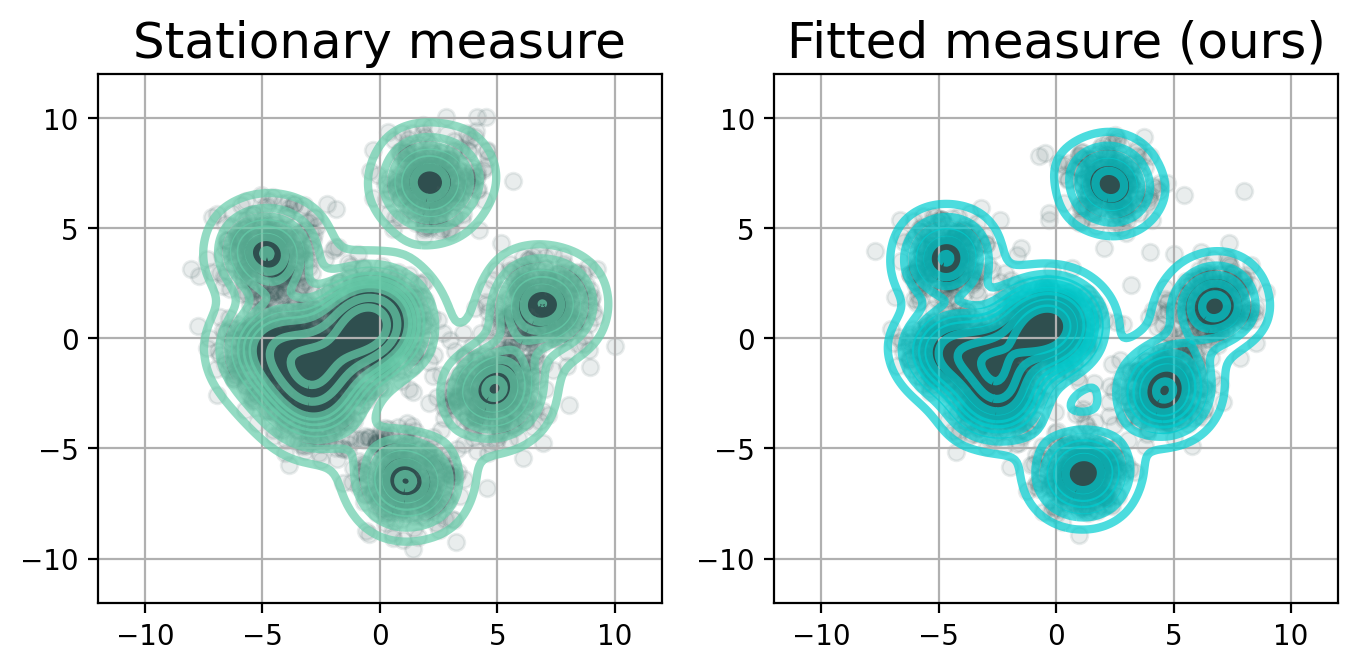}
        \caption{Dimension $n=128$}
    \end{subfigure}
    \caption{Comparison between the target GMM and fitted measure of generated samples by our method. Samples are projected onto 2D plane by performing PCA. 
    }
    \label{fig:spherical_gmm}
\end{figure}

We first consider the sampling problem to sample from a target distribution $Q$. Note that $Q$ doesn't have to be normalized.
To this end, we consider the Wasserstein gradient flow with objective function $\mF(P)= \mD(P\|Q)$, that is, the KL divergence between distributions $P$ and $Q$.
When this objective is minimized, $P\propto Q$.
In our experiments, we consider
the Gaussian mixture model (GMM) with 10 equal-weighted spherical Gaussian components. The mean of Gaussian components are randomly uniformly sampled inside a cube. The step size is set to be $a=0.1$ and the initial measure is a spherical Gaussian $\mN(0, 16I_n)$. In Figure \ref{fig:spherical_gmm}, we show our generated samples are in concordance with the target measure.

\begin{figure}[h]
    \centering
    \includegraphics[width=0.9\linewidth]{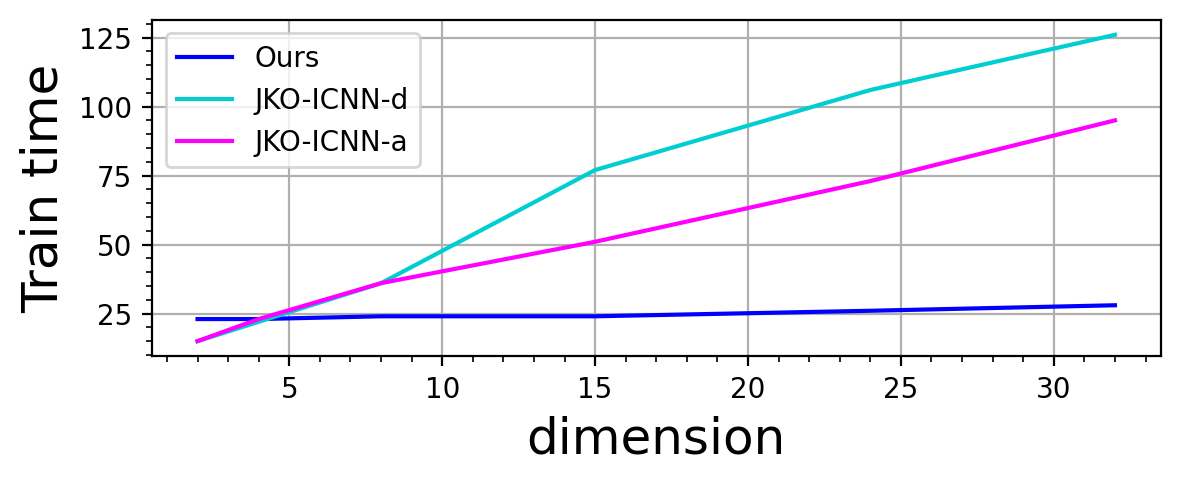}
    \caption{Averaged training time \textbf{(in minutes)} of 
    40 JKO steps for sampling from GMM. 
    }
    \label{fig:gmm_time}
\end{figure}
\begin{figure}[h]
    \centering
    \begin{subfigure}{0.23\textwidth}
        \includegraphics[width=1\linewidth]{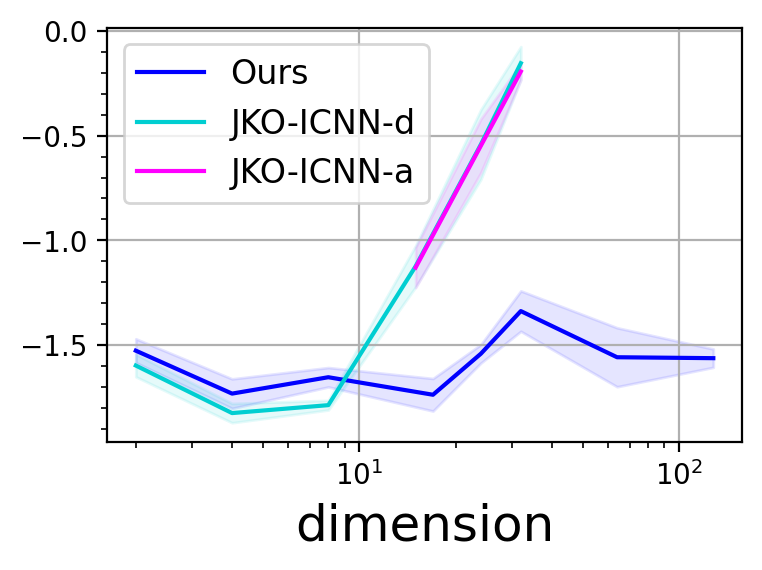}
        \caption{$\log_{10}$KSD}
    \end{subfigure}
    \begin{subfigure}{0.23\textwidth}
        \includegraphics[width=1\linewidth]{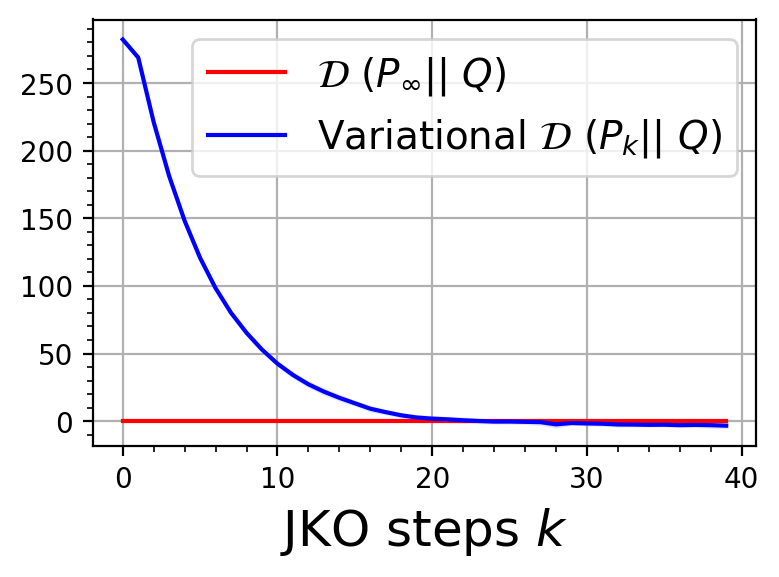} 
        \caption{Objective functional}
    \end{subfigure}
    \caption{(a) We perform experiments in $n=2,4,8,15,24,32$ for all methods and additionally $n=64,128$ for our method. With the constraint of similar training time,
    our method gives smaller error in high dimension. (b) With the variational formula, we use only samples to estimate the objective functional $\mD(P_k \| Q)$ in dimension $n=64$. It converges to the ideal objective minimum $\mD(P_\infty \| Q)=0$.
    }
    \label{fig:gmm_curve}
\end{figure}

In Figure \ref{fig:gmm_time}, we plot the averaged training time of 5 runs for all compared methods. Note that we fix the number of conjugate descent steps to be at most 10 when approximating $\log \det \nabla^2 \varphi$ in JKO-ICNN-a. That's why JKO-ICNN-d and JKO-ICNN-a have quite similar training time when $n<10$.
    
    To investigate the performance under the constraint of similar training time, we perform 40 JKO steps with our method and the same for JKO-ICNN methods except for $n \geq 15$, where we
    only let them run for $20, 15, 12$ JKO steps for $n=15,24,32$ respectively. In doing so, one can verify the training time of our method and JKO-ICNN is roughly consistent.
    We only report the accuracy results of JKO-ICNN-d for $n < 10$ in Figure \ref{fig:gmm_curve} since it's prone to give higher accuracy than JKO-ICNN-a considering nearly the same training time in low dimension. We select Kernalized Stein Divergence (KSD) \citep{liu2016kernelized} as the error criteria because it only requires samples to estimate the divergence, which is useful in the sampling task.
\subsection{Ornstein-Uhlenbeck Process} \label{sec:ou}
\begin{figure}[h]
    \vspace{-0.2cm}
    \centering
    \begin{subfigure}{0.23\textwidth}
        \includegraphics[width=1\linewidth]{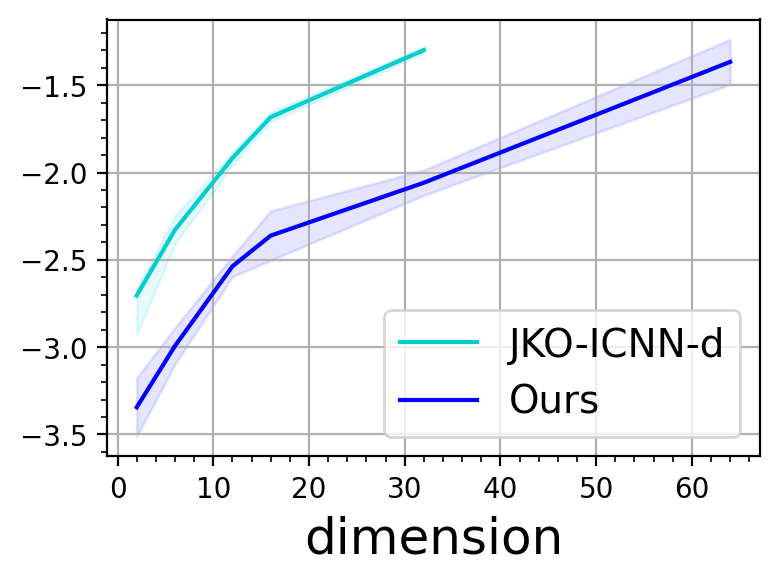}
        \caption{$\log_{10}$SymKL}
    \end{subfigure}
    \begin{subfigure}{0.23\textwidth}
        \includegraphics[width=1\linewidth]{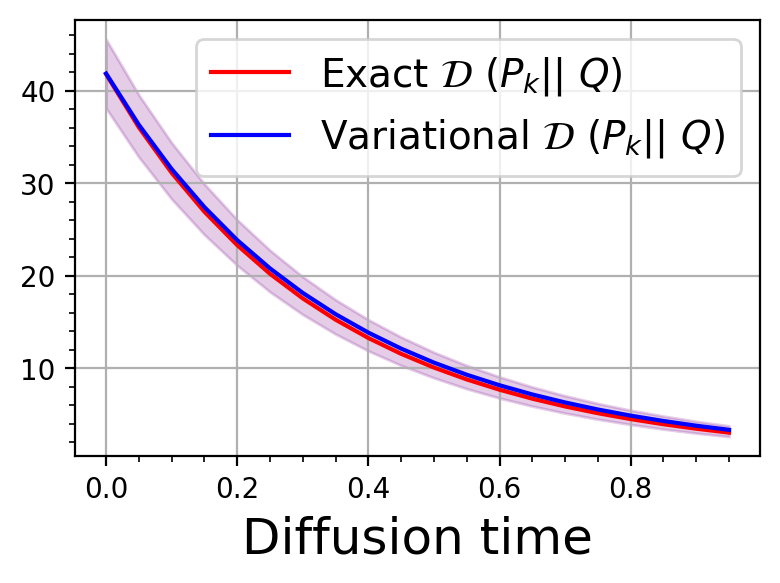}
        \caption{Objective functional}
    \end{subfigure}    
    \caption{(a): 
    We repeat the experiments for 15 times and compare the SymKL \citep{MokKorLiBur21}
    between estimated distribution and the ground truth at $k=18$ in OU process. (b): We show the comparison between our estimated $\mD(P_k\| Q)$ and the ground truth in dimension $n=64$. They align with each other pretty well.}
    \label{fig:ou}
\end{figure}
We study the performance of our method in modeling the Ornstein-Uhlenbeck Process as dimension grows. The gradient flow is affiliated with the free energy \eqref{eq:kl}, where $Q=e^{(x-b)^\tT A (x-b)/2}$ with a positive definite matrix $A \in \mR^n \times \mR^n$ and $b\in\mR^n$. Given an initial Gaussian distribution $\mN(0,I_n)$, the gradient flow at each time $t$ is a Gaussian distribution $P_t$ with mean vector 
$$\mu_t=(I_n - e^{-At})b$$
and covariance~\citep{vatiwutipong2019alternative}
$$\Sigma_t= A^{-1}(I_n - e^{-2At}) + e^{-2At}.$$ We choose JKO step size $a=0.05$. We only present JKO-ICNN-d accuracy results because JKO-ICNN-a has the similar or slightly worse performance.

{There could be several reasons why we have better performance. 1) The proposed distribution $\mu$ is Gaussian, which is consistent with $P_t$ for any $t$. This is beneficial for the inner maximization to find a precise $h$. 2) Parameterizing $T$ as a neural network instead of gradient of ICNN is handier for optimization in this toy example. 
}

We also compare the training time per every two JKO steps with JKO-ICNN method. The computation time  for JKO-ICNN-d is around 25$s$ when $n=2$ and increases to 105$s$ when $n=32$. JKO-ICNN-a has slightly better scalability, which increases from 25$s$ to 95$s$.
Our method's training time remains at $22s \pm 5s$ for all the dimensions $n=2 \sim 32$. This is due to the fact that we fix the neural network size for both methods and our method's computation complexity does not depend on the dimension.

\subsection{Bayesian Logistic Regression} \label{sec:bayesian}
\begin{table}[]
    \centering
    \caption{Bayesian logistic regression accuracy and log-likelihood results.} \label{tab: bayesian}
  {\setlength{\tabcolsep}{4pt}
    \begin{tabular}{ccccccc}
        \hline                        & \multicolumn{2}{c}{ { Accuracy }} & \multicolumn{2}{c}{ { Log-Likelihood }}                                \\
        \multirow{-2}{*}{ {Dataset }} & Ours                              & \text{JKO-ICNN}                         & Ours     & \text { JKO-ICNN} \\
        \hline \text { covtype }      & {0.753}                           & 0.75                                    & -0.528   & {-0.515}          \\
        \text { splice }              & 0.84                              & 0.845                                   & -0.38    & -0.36             \\
        \text { waveform }            & {0.785}                           & 0.78                                    & {-0.455} & -0.485            \\
        \text { twonorm }             & {0.982}                           & 0.98                                    & {-0.056} & -0.059            \\
        \text { ringnorm }            & 0.73                              & {0.74}                                  & -0.5     & -0.5              \\
        \text { german }              & {0.67}                            & {0.67}                                  & {-0.59}  & -0.6              \\
        \text { image }               & {0.866}                           & 0.82                                    & {-0.394} & -0.43             \\
        \text { diabetis }            & {0.786}                           & 0.775                                   & {-0.45}  & {-0.45}           \\
        \text { banana }              & {0.55}                            & {0.55}                                  & -0.69    & -0.69             \\
        \hline
    \end{tabular}
    }
\end{table}
To evaluate our method on a real-world datast, we consider the bayesian logistic regression task with the same setting in \citet{gershman2012nonparametric}.
Given a dataset $\mL=\{l_1,\ldots,l_S \}$, a model with parameters $x\in \mR^n$ and the prior distribution $p_0(x)$, our target is to sample from the posterior distribution $$p(x|\mL) \propto p_0(x) p(\mL|x) = p_0(x) \cdot \prod_{s=1}^S p(l_s|x).$$
To this end, we let the target distribution $Q(x)=p_0(x) p(\mL|x)$ and choose $\mF(P)$ equal to $\mD(P\|Q)$.
The parameter $x $ takes the form of 
{$[\omega,\log \alpha]$}, where $\omega \in \mR^{n-1}$ is the regression weights with the prior $p_0(\omega | \alpha) = \mN(\omega, \alpha^{-1})$. $\alpha$ is a scalar with the prior $p_0(\alpha)= \text{Gamma} (\alpha | 1,0.01)$.
We test on 8 relatively small datasets ($S \leq 7400$) from \citet{mika1999fisher} and one large Covertype dataset\footnote{\url{https://www.csie.ntu.edu.tw/~cjlin/libsvmtools/datasets/binary.html}} ($S=0.58$M). The dataset is randomly split into training dataset and test dataset according to the ratio 4:1. The number of features scales from 2 to 60.
From Table \ref{tab: bayesian}, we can tell that our method achieves a comparable performance as the other.
The results of JKO-ICNN-d are adapted from \citet[Table 1]{MokKorLiBur21}. 
We present the datasets properties and comparison with another popular sampling method SVGD \citep{svgd} in Table \ref{tab: full bayesian} in the Appendix.

\subsection{Porous media equation} \label{sec:porous}
\begin{figure}[h]
    \centering
    \includegraphics[width=0.9\linewidth]{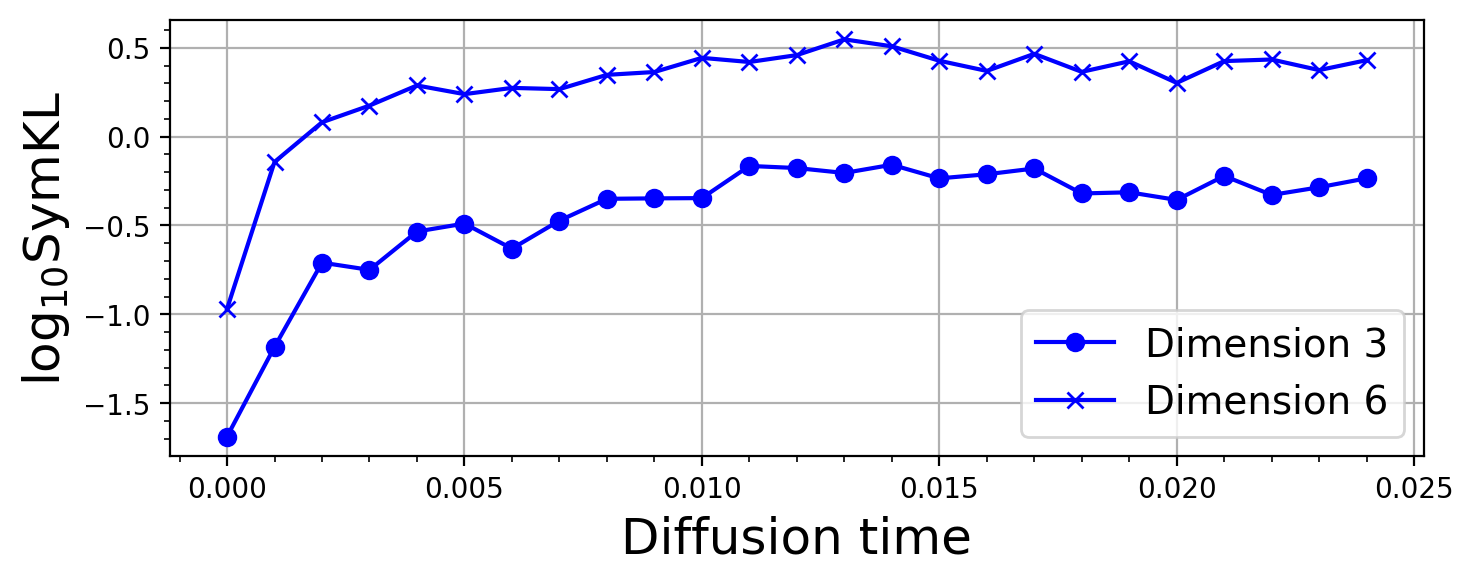}
    \caption{SymKL with respect to the Barenblatt profile ground truth in 50 JKO steps.
    }
    \label{fig:porous_highd}
\end{figure}
\begin{figure}[h]
    \centering
    \begin{subfigure}{0.23\textwidth}
        \includegraphics[width=1\linewidth]{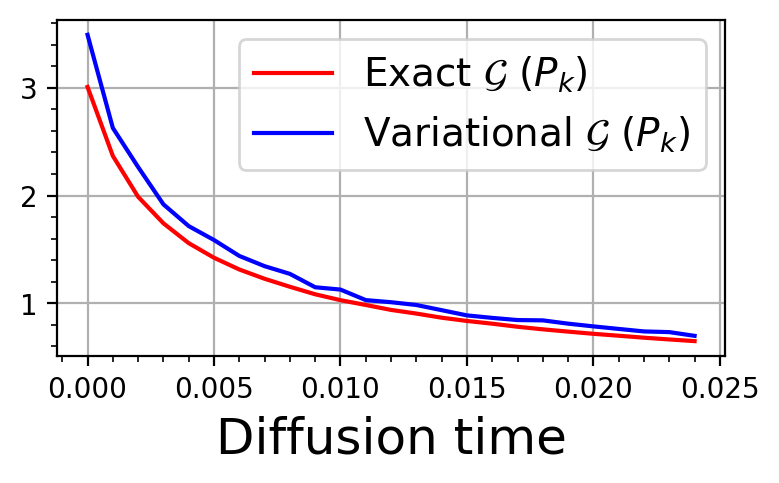}
        \caption{Dimension $n=3$}
    \end{subfigure}
    \begin{subfigure}{0.23\textwidth}
        \includegraphics[width=1\linewidth]{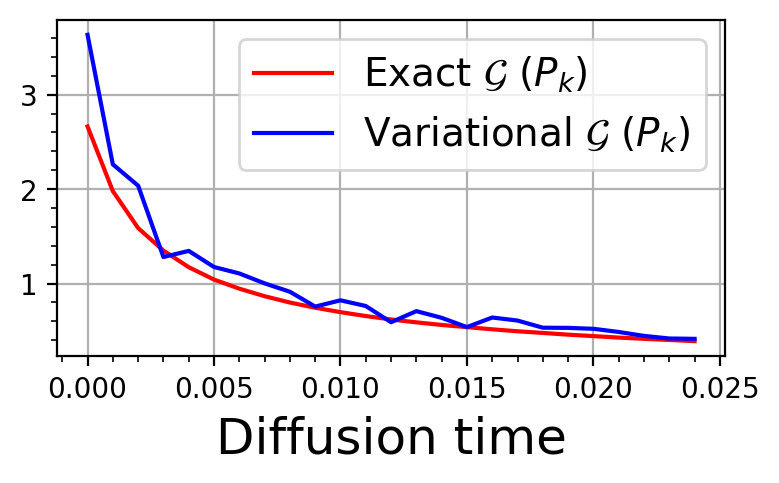}
        \caption{Dimension $n=6$}
    \end{subfigure}
    \caption{We use variational formula to calculate the objective functional $\mG(P)$ with samples and compare it with ground truth.
    }
    \label{fig:porous_var}
\end{figure}
We next consider the porous media equation with only diffusion: $\partial_t P = \Delta P^m$. This is the Wasserstein gradient flow associated with the energy function $\mF (P)=
\mG(P)$.  A representative closed-form solution of the porous media equation is the Barenblatt profile \citep{Gi52, Vaz07}
\[
     P(t,x)=\left(t+t_{0}\right)^{-\alpha}\left(C\!-\!\beta \|x-x_{0} \|^{2}\left(t+t_{0}\right)^{\frac{-2 \alpha}{n}}\right)_{+}^{\frac{1}{m-1}},
\]
where $\alpha=\frac{n}{n(m-1)+2}$, $\beta=\frac{(m-1) \alpha}{2 m n}$,
$t_0>0$ is the starting time, and $C>0$ is a free parameter.
In the experiments, we set $m=2$, the stepsize for the JKO scheme to be $a=0.0005$ and the initial time to be $t_0=0.001$.  We parametrize the transport map $T$ as the gradient of an ICNN and thus we can evaluate the density following Section \ref{sec:density}.
From Figure \ref{fig:porous_highd}, we observe that our method can give stable simulation results, where the error is controlled in a small region as diffusion time increases.

\subsection{Gradient flow on images} \label{sec:image}
\begin{figure}[h]
    \centering
    \begin{subfigure}{0.26\textwidth}
        \includegraphics[width=0.97\linewidth]{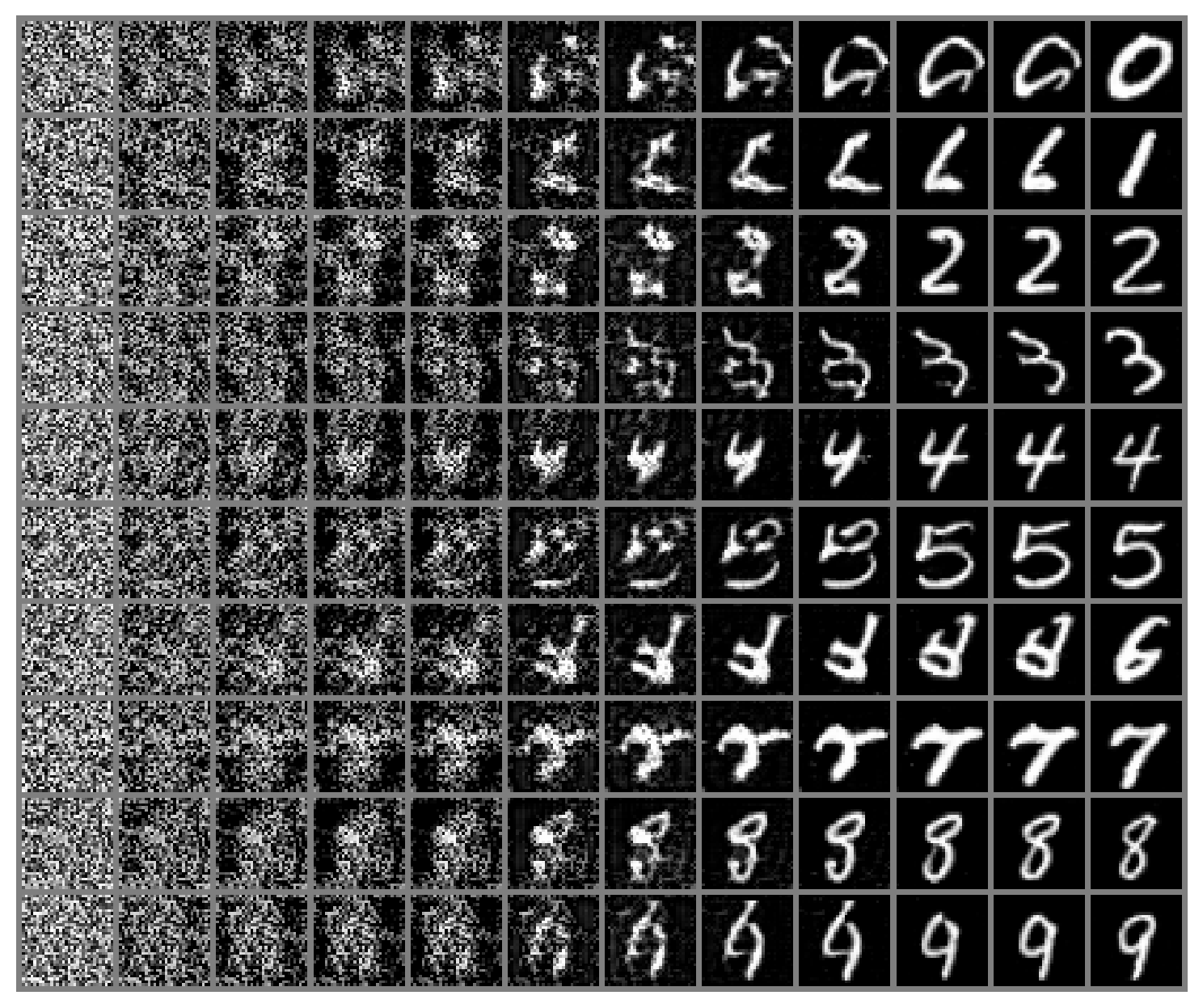}
    \end{subfigure}
    \begin{subfigure}{0.21\textwidth}
        \includegraphics[width=1\linewidth]{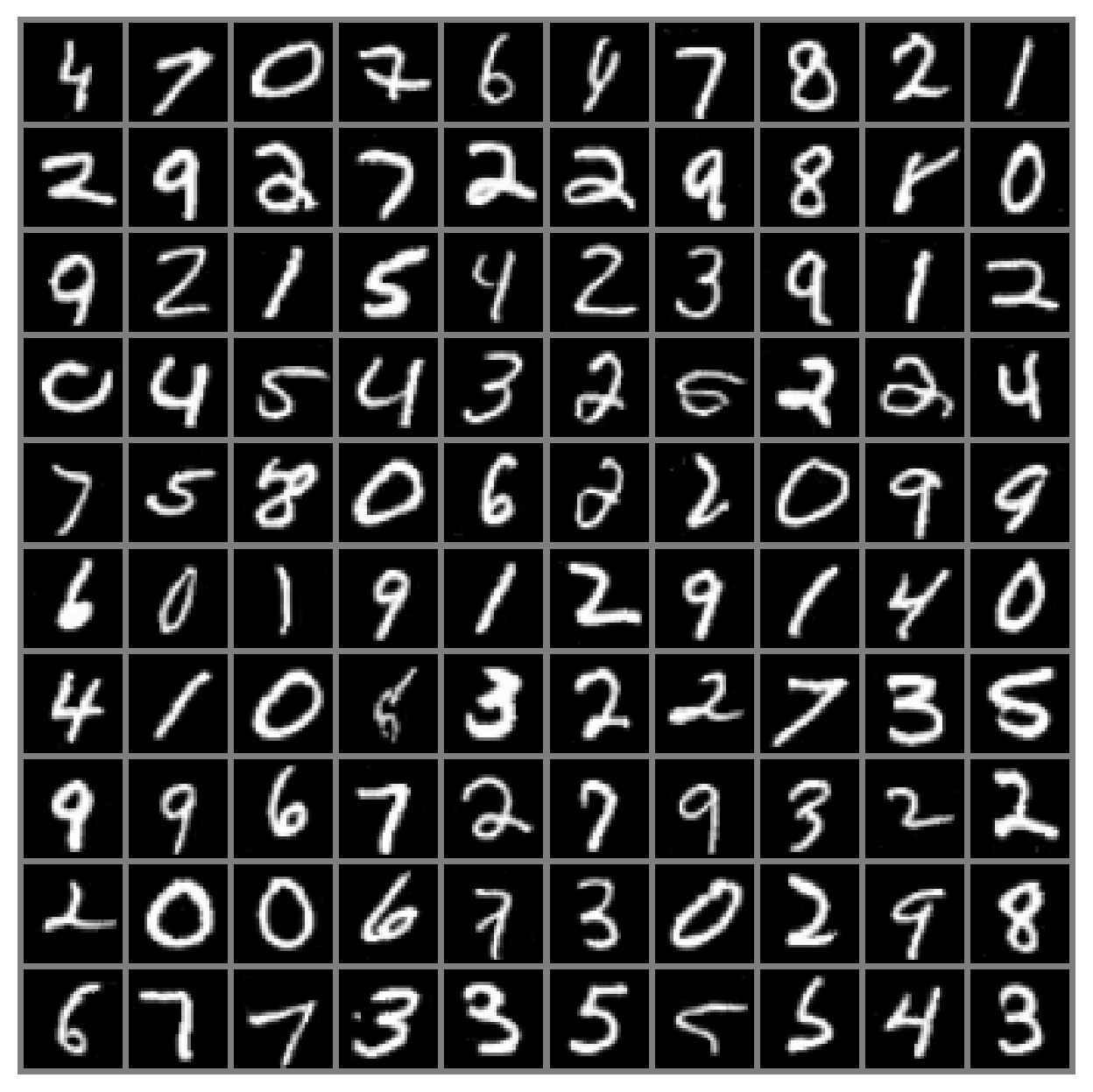}
    \end{subfigure}    
    \begin{subfigure}{0.26\textwidth}
        \includegraphics[width=1\linewidth]{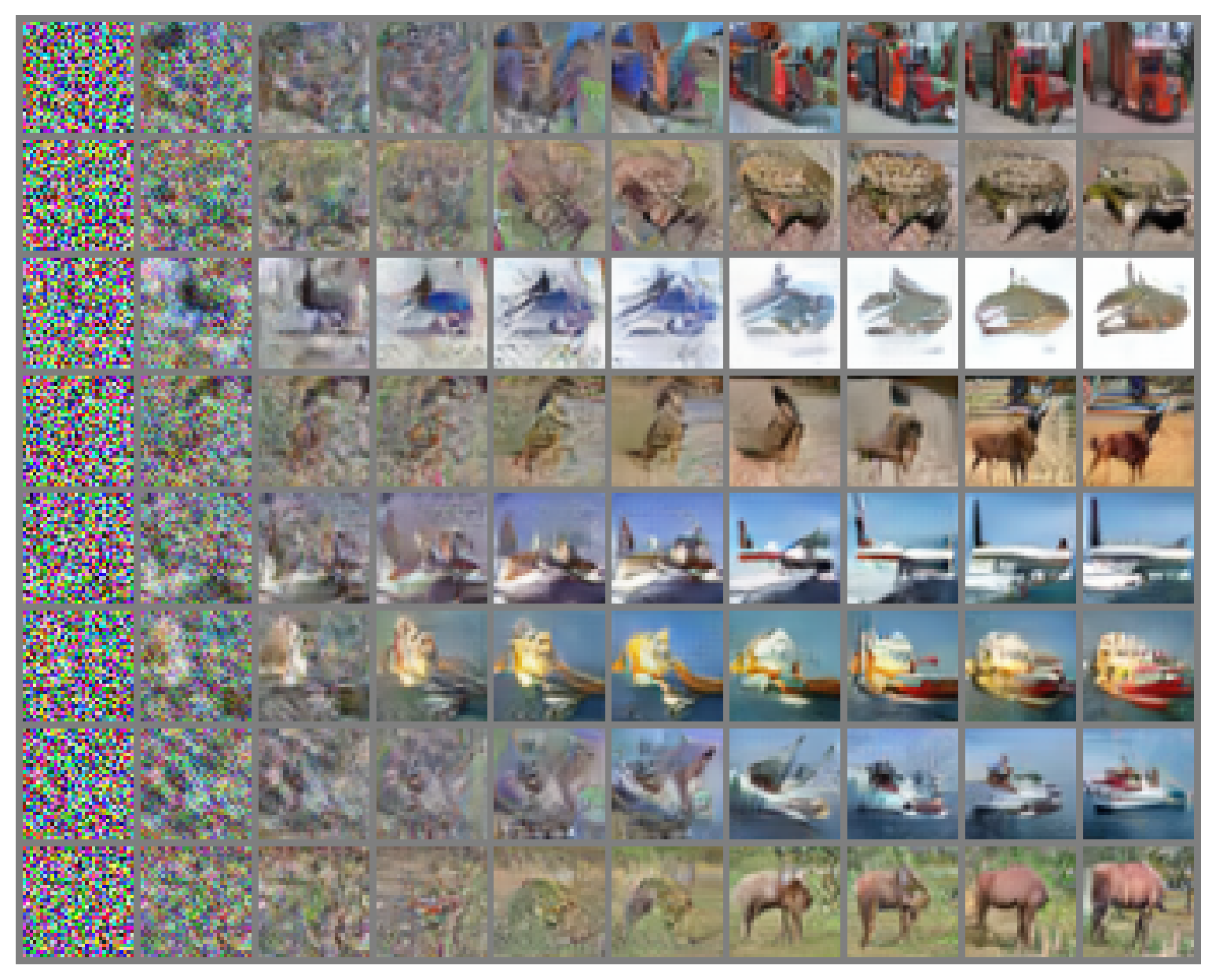}
        \caption{Trajectory}
    \end{subfigure}
    \begin{subfigure}{0.21\textwidth}
        \includegraphics[width=1\linewidth]{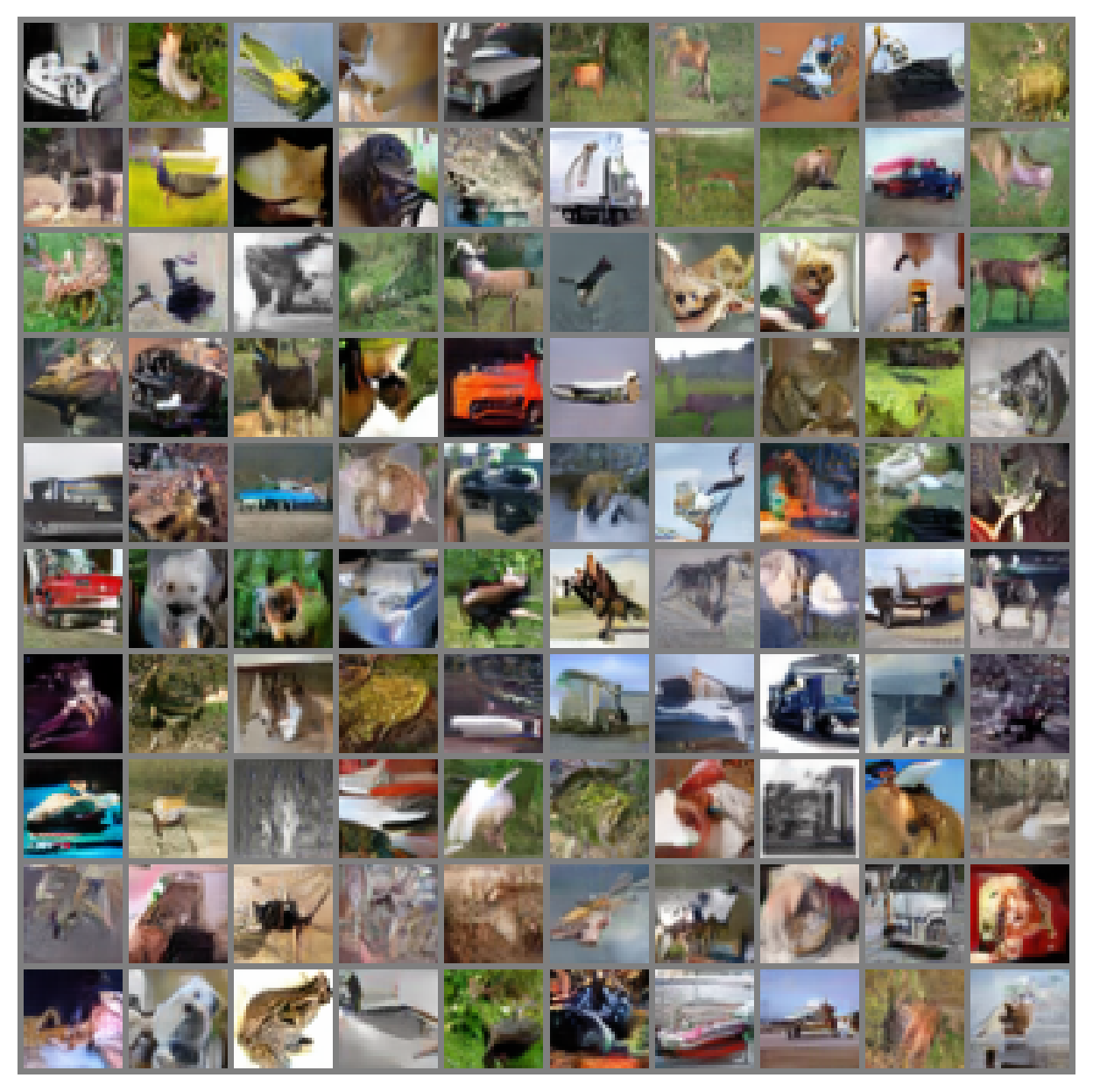}
        \caption{Uncurated samples}
    \end{subfigure}
    \caption{With Wasserstein gradient flow scheme, we visualize (a): trajectories of the generated samples from JKO-Flow and (b): 100 uncurated samples from $P_K$.
    }
    \label{fig:image}
\end{figure}
\fan{In this section, we illustrate the scalability of our algorithm to high-dimensional setting by applying} our scheme on real image datasets, 
{where only samples from $Q$ are accessible.}
With the variational formula \eqref{eq:js}, Algorithm ~\ref{algo:flow} can be adapted to model gradient flow in image space. Specifically, we choose $\mF(P)$ to be JSD$(P\|Q)$ and $P_0=\mN(0,I_n)$. We name the resulted model \textit{JKO-Flow}. 
Note JKO-Flow model specializes to GAN~\cite{goodfellow2014generative} when $a \to \infty$ and $K=1$. 
{Thanks to the additional Wasserstein loss regularization,} JKO-Flow enjoys stable training and suffer less from mode collapsing empirically. 
We evaluate JKO-Flow on popular MNIST \citep{lecun1998gradient} and CIFAR10 \citep{krizhevsky2009learning} datasets. Figure~\ref{fig:image} shows samples and their trajectories starting from $P_0$ to $P_K$ and demonstrates JKO-Flow can approximate Wasserstein gradient flow in image space empirically. 
To further quantify the performance of JKO-Flow, we measure discrepancy between $P_K$ and real distribution with the popular sample metric, Fenchel Inception Distance~\citep{heusel2017gans} in Table~\ref{table:fid}. \fan{We also compare our method with normalizing flow (NF), which also consists of a sequence of forward mapping. However, the invertible property of NF either requires heavy calculations (e.g. evaluating matrix determinant or solving Neural ODE) or special network structures that limit the the expressiveness of NNs.}
We include more comparison and experiments details in Section~\ref{app:img-exp}.

\begin{table}[]
    \caption{Results of Gradient flow (GF) based methods, GAN methods and normalizing flow (NF) on unconditional CIFAR10 dataset.}
    \begin{center}
        \begin{small}
            \begin{tabular}{c|cc}
                \toprule
              & Method  & FID score $\downarrow$ \\
                \toprule
              {NF } &   GLOW~\citep{kingma2018glow}    & \textbf{45.99} \\       
\hline \rule{0pt}{2ex}              
           & VGrow \citep{gao2019deep}    & 28.8 \\
              \multirow{-2}{*}{ {GF }}  &       JKO-Flow    & \textbf{23.1} \\
             \hline \rule{0pt}{2ex}
             &   WGAN-GP \citep{arbel2018gradient}    & 31.1 \\                
   
              \multirow{-2}{*}{ {GANs }} &   SN-GAN \citep{miyato2018spectral}    & \textbf{21.7} \\
                \bottomrule
            \end{tabular}
        \end{small}
    \end{center}
    \vskip -0.1in
    \label{table:fid}
\end{table}


\section{Conclusion}
In this paper, we presented a numerical procedure to implement the Wasserstein gradient flow for objective functions in the form of $f$-divergence. Our procedure is based on applying the JKO scheme on a variational formulation of the $f$-divergence. Each step involves solving a min-max stochastic optimization problem for a transport map and a dual function that are parameterized by neural networks. We demonstrated the scalability of our approach to high-dimensional problems through numerical experiments on Gaussian mixture models and real datasets including MNIST and CIFAR10. We also provided preliminary theoretical results regarding the variational objective function. The results show that minimizing the variational objective is meaningful and serve as starting point for future research. 
Our method can also be adapted to Crank-Nicolson type scheme, which enjoys a faster convergence \citep{CarCraWanWei21} in step size $a$ than the classical JKO scheme (see Section \ref{sec:c-n}). 
\fan{One restriction of our method is that it is only applicable to $f$-divergence, thus 
a possible direction for future research is to extend the variational formulation beyond $f$-divergence.
Another limitation is that the min-max training is both theoretically and numerically more challenging than a single minimization. 
}

\vspace{-0.3cm}
\section*{Acknowledgement}
The authors would like to thank the anonymous reviewers for useful comments. JF, QZ, and YC are supported in part by grants NSF CAREER ECCS-1942523, NSF ECCS-1901599, and NSF CCF-2008513.

\newpage
\bibliography{reference}
\bibliographystyle{icml2022}

\newpage
\appendix
\onecolumn


\fan{The appendix is structured as follows.
In Section \ref{sec:proof}, we provide the proofs of Corollaries in Section \ref{sec:var_form} and the theoretical results in Section \ref{sec:theory}. In Section \ref{sec:c-n}, we give a Crank-Nicolson-typed extension of our method for a faster convergence with respect to the step size $a$. In Section \ref{sec:aggreg}, we consider the case where the target functional $\mF(P)$ involves the interaction energy, and propose to use forward-backward scheme to solve the Wasserstein GF. In Section \ref{sec:density}, for the sake of completeness, we discuss how to evaluate the probability density of each JKO step $P_k$. In Section \ref{sec:addition}, we provide additional experimental results and discussions, such as the computational time. In Section \ref{sec:params}, we provide the training details of experiments other than image generation. In Section \ref{sec:image}, we provide the training details and discussions of image generation experiment.
}

\section{Proofs}\label{sec:proof}
\subsection{Proof of variational formulas in Section \ref{sec:var_form}} \label{sec:detail_var}
\subsubsection{KL divergence}
The KL divergence is the special instance of the $f$-divergence obtained by
replacing $f$ with $f_1(x) = x \log x$ in \eqref{eq:f-divergence}
$$D_{f_1}(P\|Q )=
    \mE_Q \left[ \frac{P}{Q}\log \frac{P}{Q}\right] =  \mE_P \left[\log \frac{P}{Q}\right] ,
$$
which, according to~\eqref{eq:f-divergence-dual}, admits the variational formulation
\begin{align}\label{eq:kl_f-divergence}
    D_{f_1}(P\|Q )=1+\sup_h \mE_P \left[ h(X)\right]- \mE_Q \left[e^{h(Z)}\right]
\end{align}
where the convex conjugate $f^*_1(y)=e^{y-1}$ 
\fan{and a change of variable $h \rightarrow h-1$ are }
used.

The variational formulation can be approximated in terms of samples from $P$ and $Q$. For the case where we have only access to un-normalized density of $Q$, which is the case for the sampling problem, we use the following change of variable:
$h\to  \log (h) + \log(\mu)-\log(Q)$ where $\mu$ is a user designed distribution which is easy to sample from. Under such a change of variable, the variational formulation reads
\begin{equation*}
    D_{f_1}(P\|Q )=1+\sup_h \mathbb{E}_{P}\left[  \log h(X) + \log\frac{\mu (X)}{Q(X)} \right] - \mE_\mu \left[{h(Z)}\right].
\end{equation*}
Note that the optimal function $h$ is equal to the ratio between the densities of ${T\sharp P_k}$ and ${\mu}$.
 Using this variational form in the JKO scheme~\eqref{eq:JKO-T} yields $P_{k+1} =T_k\sharp P_k$ and
\begin{align} \label{eq:kl_jko}
    T_k     =
     \argmin_{T} \max_h \mE_{P_k}\left[ \frac{\|X-T(X)\|^2}{2a} +  \log h(T(X)) + \log\frac{\mu (T(X))}{Q(T(X))} \right] - \mE_{\mu} \left[{h(Z)}\right].
\end{align}
\fan{Based on particle approximation, the implementable JKO is
\begin{align} \label{eq:kl_jko_sample}
T_k = \argmin_{T} \max_h 
\frac{1}{N} \sum_{i=1}^N
\left[\frac{\|X_i^{(k)} -T(X_i^{(k)})\|^2}{2a} +  \log h(T(X_i^{(k)})) + \log\frac{\mu (T(X_i^{(k)}))}{Q(T(X_i^{(k)}))} \right] - \mE_{\mu} \left[{h(Z)}\right].
\end{align}
}

\begin{remark}
    The Donsker-Varadhan formula
    \begin{align*}
        \mD(P\|Q )=\sup_h \mE_P \left[ h(X)\right]- \log \mE_Q \left[e^{h(Z)}\right]
    \end{align*}
    is another variational representation of KL divergence and it's a stronger than \eqref{eq:kl_f-divergence} because it's a upper bound of \eqref{eq:kl_f-divergence} for any fixed $h$.
    However, we cannot get an unbiased estimation of the objective using samples.
\end{remark}

\subsubsection{{Generalized entropy}}
The generalized entropy can be also represented as $f$-divergence. In particular, let  $f_2(x)=\frac{1}{m-1} (x^m-x)$ and let $Q$ be the uniform distribution on a set which is the superset of the support of density $P(x)$ and has volume $\Omega$. Then
\begin{align*}
    D_{f_2}(P\| Q)=\frac{\Omega^{m-1}}{m-1} \int P^m(x)dx -\frac{1}{m-1}.
\end{align*}
As a result, the generalized entropy can be expressed in terms of $f$-divergence according to
\begin{equation*}
    \mG(P)=\frac{1}{m-1} \int P^m(x)dx = \frac{1}{\Omega^{m-1}}D_{f_2}(P\| Q)+ \frac{1}{\Omega^{m-1}(m-1)}.
\end{equation*}
Upon using the variational representation of the $f$-divergence with $$f_2^*(y)= \left( \frac{(m-1)y+1}{m}\right)^{\frac{m}{m-1}},$$ the generalized entropy admits the following variational formulation
\begin{align*}
    \mG(P)
    =\sup_h \frac{1}{\Omega^{m-1}} \left( \mE_P [h(X)]-\mE_Q \left[\left( \frac{(m-1)h(Z)+1}{m}\right)^{\frac{m}{m-1}} \right] \right)
    + \frac{1}{\Omega^{m-1}(m-1)}.
\end{align*}
In practice, we find it numerically useful to let
$h=\frac{1}{m-1} \left[ m \left(\hat{h} \right)^{m-1}-1 \right]$ so that
\begin{align}
    \mG(P) =\frac{1}{\Omega^{m-1}} \sup_{\hat{h}} \left( \mE_{P_k} \left[\frac{m}{m-1} \hat{h}^{m-1}(X)  \right]-\mE_Q \left[\hat{h}^m(Z)  \right] \right).
\end{align}
With such a change of variable, the optimal function $\hat{h}=T\sharp P_k/Q$.
Using this in the JKO scheme yields $ P_{k+1}  =T_k \sharp P_k$, and
\begin{align*}
      T_k     =\argmin_{T} \max_h
    \frac{1}{2a} \mE_{P_k}\|X-T(X)\|^2           + \frac{1}{\Omega^{m-1}} \left( \mE_{P_k} \left[\frac{m}{m-1} h^{m-1}(X)  \right]-\mE_Q \left[h^m(Z)  \right] \right).
\end{align*}

\subsubsection{Jensen-Shannon divergence}
Jensen-Shannon divergence has been widely studied in GAN literature \citep{nowozin2016f}. The variational formula follows that  $f(x) =-(x+1)\log((1+x)/2) +x\log x $ and $f^*(y)= - \log(2-\exp(y))$. 
Plugging in the variational formula in the JKO scheme gives 
\begin{align*}
      T_k     =\argmin_{T} \max_h
    \frac{1}{2a} \mE_{P_k}\|X-T(X)\|^2           +  \mE_{P_k} \left[\log(1-h(X)) \right] + \mE_Q \left[\log {h}(Z)  \right].
\end{align*}





\subsection{Proof of Propostion~\ref{lem:DfH}}
We present the proof of Propostion~\ref{lem:DfH}. Let us define $J(h) := \int hd P - \int f^*(h)dQ$. 
\begin{enumerate}
    \item The proof follows from 
    \begin{equation*}
        D_f^{\mathcal H}(P,Q) = \sup_{h\in \mathcal H} J(h) \geq \sup_{c\in \mathbb{R}} J(c) = \sup_{c \in \mathbb{R}} \{c - f^*(c)\} = f(1) =  0 
    \end{equation*}
    where the last identity follows from the assumption that $f(1)=0$.
    \item The direction ($\Leftarrow$) follows because 
    \begin{equation*}
        J(h) \leq \int hd P - \int h dQ  = 0,\quad \forall h \in \mathcal H
    \end{equation*}
    where $f^*(y)= \sup_{x}\{xy - f(x)\} \geq y1-f(1) = y$ is used. As a result, $D_f^{\mathcal H}(p\|Q) = \sup_{h \in \mathcal H} J(h) \leq 0$. Using part (1), this is only possible when $D_f^{\mathcal H}(P\|Q)=0$. 
    
    To show the other direction ($\Rightarrow$), for all $h \in \mathcal H$ , define $g(\lambda):= J(f'(1)+\lambda h)$ where $\lambda\in \mathbb{R}$. The function $g(\lambda)$ attains its maximum at $\lambda =0$ because $g(\lambda) = J(f'(1)+\lambda h) \leq \sup_{h\in \mathcal H} J(h) = D_f^{\mathcal H}(P\|Q)=0$ and $g(0)=J(f'(1))=f'(1)-f^*(f'(1))=f(1)=0$ by Fenchel identity. Therefore, the first-order optimality condition $g'(0)=0$ must hold. The result follows because 
    \begin{equation*}
      g'(0) = \int h d P - \int h {f^*}' (f'(1)) d Q = \int h d P - \int h  d Q 
    \end{equation*}
    \item Let us define $g_h(\lambda) := J(f'(1)+ \frac{\lambda h}{\|h\|_{2,Q}})$. The first and the second derivatives of $g_h(\lambda)$ with respect to $\lambda$ are:
    \begin{align*}
        g'_h(\lambda) &= \int \frac{h}{\|h\|} dP - \int \frac{h}{\|h\|} {f^*}'(f'(1)+\frac{\lambda h}{\|h\|})dQ\\
        g''_h(\lambda) &=- \int \frac{h^2}{\|h\|^2} {f^*}''(f'(1)+\frac{\lambda h}{\|h\|})dQ  
    \end{align*}
    By assumption on $f$, the convex conjugate $f^*$ is strongly convex with constant $\frac{1}{L}$ and smooth with constant $\frac{1}{\alpha}$. Therefore, $\frac{1}{L}\leq {f^*}''(y)\leq \frac{1}{\alpha}$. As a result, $\frac{1}{L}\leq -g''_h(\lambda)\leq  \frac{1}{\alpha}$ where we used $\|h\|^2 = \int h^2 dQ$. Therefore, $g_h(\lambda)$ is strongly concave and smooth and satisfies the inequalities: 
    \begin{align*}
         \frac{\alpha}{2} g'_h(0)^2 \leq \sup_{\lambda}g_h(\lambda) - g_h(0)\leq \frac{L}{2} g'_h(0)^2
    \end{align*}
    Upon using $g_h(0) = J(f'(1))=0$ and taking the sup over $h\in \mathcal H$ of all sides, 
        \begin{align*}
         \frac{\alpha}{2} \sup_{h\in \mathcal H}g'_h(0)^2 \leq  \sup_{h\in \mathcal H} \sup_{\lambda}g_h(\lambda) \leq \frac{L}{2}  \sup_{h\in \mathcal H} g'_h(0)^2.
    \end{align*}
    By the assumption that for all $h \in \mathcal H$, $c + \lambda h \in \mathcal H$ for $c,\lambda \in \mathbb{R}$, 
    $$\sup_{h \in \mathcal H}\sup_{\lambda}g_h(\lambda) = \sup_{h\in \mathcal H}J(h) = D_f^{\mathcal H}(P\|Q).$$
    The result follows by noting that $\sup_{h \in \mathcal H} g'_h(0) = d_{\mathcal H}(P,Q)$. 
\end{enumerate}

\subsection{Proof of Proposition \ref{lem:low_bound}}\label{sec:low_bound_proof}
\begin{proof}
For a given $P$ and $Q$, let $h_0=f'(
\frac{dP}{dQ})$ and $\tilde{h}\in \mathcal H$ be such that $\|\tilde{h}-h_0\|_{2,Q}\leq \epsilon$. Similar to the proof of Proposition~\ref{lem:DfH}, define $J(h) = \int h dP - \int f^*(h)dQ$. Then, 
\begin{equation*}
    D_f^{\mathcal H}(P\|Q)  = \sup_{h\in \mathcal H} J(h) \geq J(\tilde{h}) = J(\tilde{h})-J(h_0) + J(h_0) = J(\tilde{h})-J(h_0) + D_f(P\|Q)
\end{equation*}
where $J(h_0)=D_f(P\|Q)$ is used in the last step. The  proof follows by showing that $J(\tilde{h}) - J(h_0)\geq -\frac{1}{2\alpha}\|\tilde{h}-h_0\|_{2,Q}^2$. In order to show this, note that $f^*$ is $\frac{1}{\alpha}$ smooth because $f$ is $\alpha$ strongly convex. Then, 
\begin{equation*}
    f^*(\tilde{h}(x)) -f^*(h_0(x)) \leq   {f^*}'(h_0(x))(\tilde{h}(x)-h_0(x)) + \frac{1}{2\alpha} |\tilde{h}(x)-h_0(x)|^2,\quad \forall x.
\end{equation*}
Taking the expectation over $Q$ and adding $\int h_0 d P - \int \tilde{h}dP$ yields,
\begin{equation*}
    J(h_0) - J(\tilde{h}) \leq   \int {f^*}'(h_0)(\tilde{h}-h_0)dQ + \int (h_0-\tilde{h})dP+  \frac{1}{2\alpha} \|\tilde{h}-h\|_{2,Q}^2. 
\end{equation*}
Then, the proof follows from ${f^*}'(h_0) ={f^*}'(f'(\frac{dP}{dQ})) = \frac{dP}{dQ}$ to cancel the first two terms.   
\end{proof}

\fan{\paragraph{Discussion on neural network function class}
Consider $\mH$ is the class of neural nets with an arbitrary depth and mild assumption on the activation function. Following the proof of Theorem 1 in ~\citet{Korotin2022NeuralOT}, we can verify that for any $\epsilon>0$, compactly supported $Q$, and function $\|h \|_{2,Q} < \infty $, there exists a neural net $\tilde{h} \in \mH$ such that $\|\tilde{h}-h\|_{2,Q}\leq \epsilon$. Indeed, let $Q$ be supported on $\mX \subset \mR^n$, and $\mX$ be compact,
by
~\citet[Proposition 7.9]{folland1999real}, the continuous functions $C^0(\mX)$ are dense in $L^2(Q)$. Further by
~\citet[Theorem 3.2]{kidger2020universal}, the neural nets in $\mH$ are dense in $C^0(\mX )$ with respect to $L^\infty$ norm, and as such with respect to $L^2$ norm. Putting these two pieces together gives neural nets are dense in $L^2(Q)$.
}
\subsection{Proof of Proposition \ref{lem:rad}}
\begin{proof}
We first introduce the following notations
\begin{align*}
J(h) & = \int h dP - \int f^*(h)dQ\\
    J_{M,N}(h) &=\int h dP^{(N)} -\int f^*(h) dQ^{(M)}, \\ 
G_P(h) &=\int h dP -\int h d P^{(N)},\\
G_Q(h) &=\int f^*(h) dQ -\int f^*(h) d Q^{(M)} .
\end{align*}
  Assume the $\sup_{h\in \mathcal H} J(h)$ is attained at $h=\bar{h}$ and $\sup_{h\in \mathcal H} J_{M,N}(h)$ is attained at $h=h_{M,N}$.
\begin{align*}
    \sup_{h\in \mathcal H} J_{M,N}(h) - \sup_{h\in \mathcal H} J(h)
    = J_{M,N}(h_{M,N}) - \sup_{h\in \mathcal H} J(h) 
     \leq J_N(h_{M,N}) - J(h_{M,N}) 
     \leq \sup_{h \in \mathcal H}\{|G_P(h)|\} + \sup_{h \in \mathcal H}\{|G_Q(h)|\}.
\end{align*}
Similarly
\begin{align*}
    \sup_{h\in \mathcal H} J(h) - \sup_{h\in \mathcal H} J_{M,N}(h) 
    = J(\bar{h}) - \sup_{h\in \mathcal H} J_{M,N}(h) 
    \leq J_{M,N}(\bar h) - J(\bar h) \leq \sup_{h \in \mathcal H}\{|G_P(h)|\} + \sup_{h \in \mathcal H}\{|G_Q(h)|\}.
\end{align*}
Therefore,
\begin{align*}
   | D_f^{\mH}(P^{(N)}\|Q^{(M)}) - D_f^{\mH}(P\|Q) | = | \sup_{h \in \mH} J_{M,N}(
   h) - \sup_{h \in \mH} J(h)| \leq \sup_{h \in \mathcal H}\{|G_P(h)|\} + \sup_{h \in \mathcal H}\{|G_Q(h)|\}.
\end{align*}
The result follows by taking the expectation and the symmetrization inequality \citep[Lemma 5.1]{wellner2005empirical} to the last two terms
\begin{equation*}
    \mathbb{E} \sup_{h \in \mathcal H}\{|G_P(h)|\} + \mathbb{E} \sup_{h \in \mathcal H}\{|G_Q(h)|\} \leq 2 \mathcal R_N(\mathcal H,P) + 2 \mathcal R_M( f^* \circ \mathcal H,Q).
\end{equation*}
\end{proof}

It's not difficult to prove the following corollary following the same logic.

\begin{corollary}
Let $P^{(N)} = \frac{1}{N}\sum_{i=1}^N \delta_{X_i}$, where $\{X_i\}_{i=1}^N$ are i.i.d samples from $P$. Then, it follows that
\begin{align*}
 \mathbb E[|D^{\mathcal H}_f(
P\|Q) - D^{\mathcal H}_f(
P^{(N)}\|Q)|] 
\leq  2\mathcal R_N(\mathcal H,P) ,
\end{align*}
where the expectation is over the samples and  $\mathcal R_N(\mathcal H,P)$ denotes the Rademacher complexity of the function class $\mathcal H$ with respect to $P$ for sample
size $N$. 
\end{corollary}


    
    

\subsection{Proof of Proposition \ref{prop:gauss}}
\begin{proof}
Suppose $P_0=\mu=\mN(0,I),Q=\mN(\eta,I)$ and $ \mF(P)$ is the KL divergence $ \mD (P\|Q),  
$  we parameterize $T_k(x) = x+\beta_k, ~~h_k(z) = \exp(\alpha_k^\top z + \gamma_k) $.
Then the closed-form solution of JKO is $P^*_k=\mN(\eta_k, I) $ where 
$$\eta_k = \eta \left(1- \frac{1}{(1+a)^k } \right). $$
Our method adopts the JKO iteration \eqref{eq:kl_jko_sample} with the variational formula \eqref{eq:kl}. Since $\mu$ is a user-defined Gaussian distribution, it is reasonable to assume $ \mE_\mu[ h(Z) ]$ can be estimated precisely. To sample from $P_k$ at the $k$-th JKO step, we sample $N$ particles from the very beginning $\{X^k_i\}_{i=1}^N \sim P_0$ with empirical mean $\eta_0^k = \frac{1}{N}\sum_{i=1}^N X_i^k$, and pushforward them through maps $T_1,\ldots, T_{k-1}$. 
We also define $\eta_{k}^K = \frac{1}{N}\sum_{i=1}^N T_{k-1} \circ \cdots \circ T_{1} ( X_i^K )$. 
Clearly, $\eta_{k}^K = \eta_{0}^K +  \sum_{j=0}^{k-1} \beta_j ~~$for $ 1\leq k \leq K, ~~ 1\leq K \leq \infty. $
Then the solutions of our method are
\begin{align}
 \beta_k   =\frac{a(\eta - \eta_{k}^{k+1})}{1+a} ,        \quad               
 \alpha_k  = \beta_k + \eta_{k}^{k+1} - \eta_{k}^{k},  \quad       
 \gamma_k  = -\alpha_k^\top \eta_{k}^{k} -\frac{\|\alpha_k\|^2}{2}.
\end{align}
Thus the mean of $P_K$ is $\widehat\eta_K = \sum_{j=0}^{K-1} \beta_K$. By standard matrix calculation, we have
$\widehat \eta_K= \eta_K - \varepsilon_N$, where $$\varepsilon_N=
\frac{a}{1+a}\sum_{j=1}^K \frac{\eta_0^j}{(1+a)^{K-j} }. $$

We also get $\eta_K^K -\eta_K = \eta_0^K -\varepsilon_N $.
Denote $\Delta_K =\frac{\|\eta\|  }{ (1+a)^{K}} $,
and $\xi_{K,N} = \mathbb E[\| \varepsilon_N  \|^2 ] =  \left(\frac{a}{1+a}\right)^2 \frac{n}{N} \sum_{j=1}^K \frac{1}{ (1+a)^{2(K-j)} } .$
By the Corollary \ref{cor:kl}, we can derive 
\begin{align}
\mD^\mH (P_K^{(N)} \| Q ) = \|\eta_K^K -\eta\|^2/2, 
\end{align}
where $\eta_K^K  $ is the mean of $P_K^{(N)}$.
Finally,
\begin{align*}
 \mathbb E[|\mD^{\mathcal H}(
P^*_K \|Q) - \mD^{\mathcal H}(
P^{(N)}_K \|Q)|] 
= &  \mathbb E[|\|\eta_K -\eta\|^2 - \|\eta_K^K  -\eta \|^2 |]/2 \\
= & \mathbb E[|\|\eta_K -\eta\|^2 - \|\eta_K^K - \eta_K + \eta_K -\eta \|^2 |]/2 \\
= & \mathbb E[|\|\eta_K^K -\eta_K \|^2/2 - (\eta_K^K -\eta_K)^\top (\eta_K -\eta) |] \\
 \le & \mathbb E[\|\eta_K^K -\eta_K \|^2/2 ] + \mathbb E[ | (\eta_K^K -\eta_K)^\top (\eta_K -\eta) |]  \\
 = & \mathbb E[\|\eta_0^K - \varepsilon_N \|^2/2 ] + \mathbb E[ | (\eta_0^K - \varepsilon_N )^\top ( \eta_K -\eta ) |] \\
 \le  & \mathbb E[\|\eta_0^K - \varepsilon_N  \|^2/2 ] + \| \eta_K -\eta\| \mathbb E[ \| \eta_0^K - \varepsilon_N  \|] \\
\le & \mathbb E[\|\eta_0^K - \varepsilon_N  \|^2/2 ] + \Delta_K \sqrt{\mathbb E[\|\eta_0^K - \varepsilon_N  \|^2 ]} \\
\le & \frac{\xi_{K,N}}{2 } + \frac{n}{2N} + \Delta_K \sqrt{\xi_{K,N} + \frac{n}{N}} .
\end{align*}

\end{proof}

\section{Extension to Crank-Nicolson scheme}\label{sec:c-n}
Consider the Crank-Nicolson inspired JKO scheme \citep{CarCraWanWei21} below
\begin{align}\label{eq:c-n}
    P_{k+1}=\argmin_{P \in \mP_{ac}(\mathbb{R}^n)} \frac{1}{2a} W_2^2 \left( P, P_k\right) +  \frac{1}{2} \mathcal{F}(P) +\frac{1}{2} \int \frac{\delta \mF}{ \delta P} (P_k) P.
\end{align}
The difficulty of implementing this scheme with neural-network based method is the easy access to the density of $P_k$. The predecessors \citet{MokKorLiBur21} and \citet{AlvSchMro21} don't have this property, while in our algorithm, $P_k \approx h_{k-1} \Gamma_{k-1} (k>1)$. This is because our optimal $h_k$ is equal to or can be transformed to the ratio between densities of $ P_{k+1}$ and $\Gamma_k$. Assume $h$ can learn to approximate $P_{k+1} /\Gamma_k$, our method can be natually extended to Crank-Nicolson inspired JKO scheme.

\section{Extension to the interaction energy functional} \label{sec:aggreg}
In this section, we consider $\mF(P)$ involves the interaction energy
\begin{align}
     & \mF(P) = \mW(P):=\int\!\int W(x-y) P(x) P(y) dx dy,       \\
     & W: \mR^n \rightarrow \mR \text{ is symmetric, i.e. } W(x)=W(-x).
\end{align}

\subsection{Forward Backward (FB) scheme} \label{sec:fb}

When $\mF(P)$ involves the interaction energy $\mW(P),$
we add an additional forward step to solve the gradient flow:
\begin{align}
    P_{k+\frac{1}{2} } & := (I- a \nabla_x( W*P_k ))\sharp P_k \label{eq:fb1}           \\
    P_{k+1}            & :=T_{k + \frac{1}{2}}\sharp P_{k+\frac{1}{2}}  ,\label{eq:fb2}
\end{align}
where $I $ is the identity map,  and $T_{k+\frac{1}{2}}$ is defined by replacing $k$ by $k+\frac{1}{2}$ in \eqref{eq:saddle}. In other words, the first gradient descent step \eqref{eq:fb1} is a forward discretization of the gradient flow and the second JKO step \eqref{eq:fb2} is a backward discretization.
$\nabla_x( W*P )$ can be written as expectation $\mE_{y \sim P}\nabla_x( W(x-y) )$, thus can also be approximated by samples. The computation complexity of step \eqref{eq:fb1} is at most $O(N^2)$ where $N$ is the total number of particles to push-forward. 
This scheme has been studied as a discretization of gradient flows and proved to have sublinear convergence to the minimizer of $\mF(P)$ under some regular assumptions \citep{SalKorLui20}.
We make this scheme practical by giving a scalable implementation of JKO. 

Since $\mW(P)$ can be equivalently written as expectation $\mE_{x,y\sim P} [W(x - y)]$,
there exists another non-forward-backward (non-FB) method , i.e., removing the first step and integrating $\mW(P)$ into a single JKO step: $P_{k+1}=T_k\sharp P_k$ and
\begin{align*}
    T_k = & \argmin_T ( \mathbb{E}_{P_k}\|X-T(X)\|^2 /2a \\
    +     & \mE_{X,Y\sim P_k} [W(T(X) - T(Y))]
    +  \sup_h \mV(T,h)
    ).
\end{align*}
In practice, we observe the FB scheme is more stable and gives more regular results however converge slower than non-FB scheme.  The detailed discussion appears in the Appendix \ref{sec:W only}, \ref{sec:non-prox agg-diff}.

\begin{remark}
    In principle, one can single out $\log (Q)$ term from \eqref{eq:kl_jko} and perform a similar forward step $P_{k+\frac{1}{2}}=(I - a(\nabla_x Q)/Q)\sharp P_{k}$ \citep{SalKorLui20}, but we don't observe improved performance of doing this in sampling task.
\end{remark}

\subsection{Simulation solutions to Aggregation equation  }\label{sec:W only}
\citet{AlvSchMro21} proposes using the neural network based JKO, i.e. the backward method, to solve \eqref{eq:interaction}. They parameterize $T$ as the gradient of the ICNN. In this section, we use two cases to compare the forward method and backward  when $\mF(P)=\mW(P) $.
This could help explain the FB and non-FB scheme performance difference later in Section \ref{sec:non-prox agg-diff}.

We study the gradient flow associated with the aggregation equation
\begin{align}\label{eq:interaction}
    \partial_t P=\nabla \cdot (P \nabla W *P), \quad W:\mR^n \rightarrow \mR.
\end{align}
The forward method is
\begin{align*}
    P_{k+1 } := (I- a \nabla_x( W*P_k ))\sharp P_k.
\end{align*}
The backward method or JKO is
\begin{align*}
    P_{k+1 } := T_k \sharp P_k, \quad  T_k = \argmin_T \left\{ \frac{1}{2a} \mathbb{E}_{P_k}[\|X-T(X)\|^2]
    +\mE_{X,Y\sim P_k} [W(T(X) - T(Y))]
    \right\}.
\end{align*}

\begin{figure}[h]
    \centering
    \begin{subfigure}{0.45\textwidth}
        \includegraphics[width=0.45\linewidth]{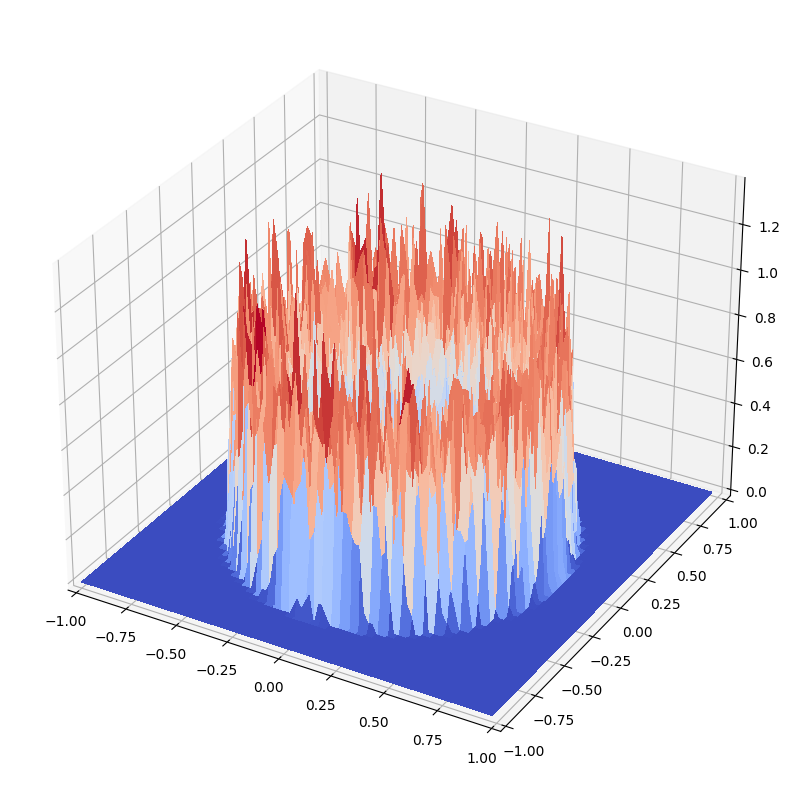}
        \includegraphics[width=0.45\linewidth]{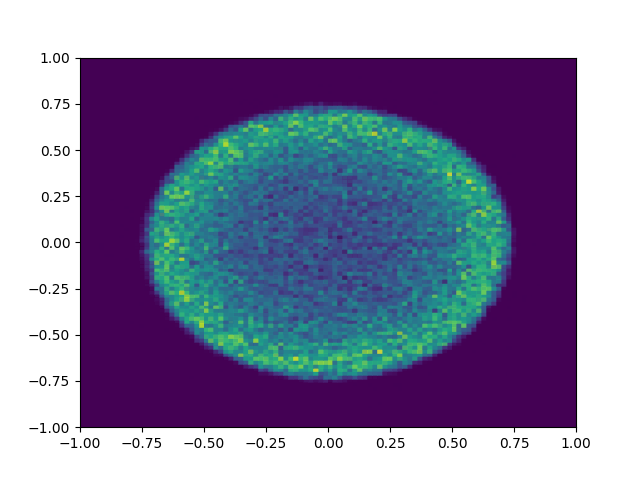}
        \caption{Forward method $k=23, t=1.15$}
    \end{subfigure}
    \begin{subfigure}{0.45\textwidth}
        \includegraphics[width=0.45\linewidth]{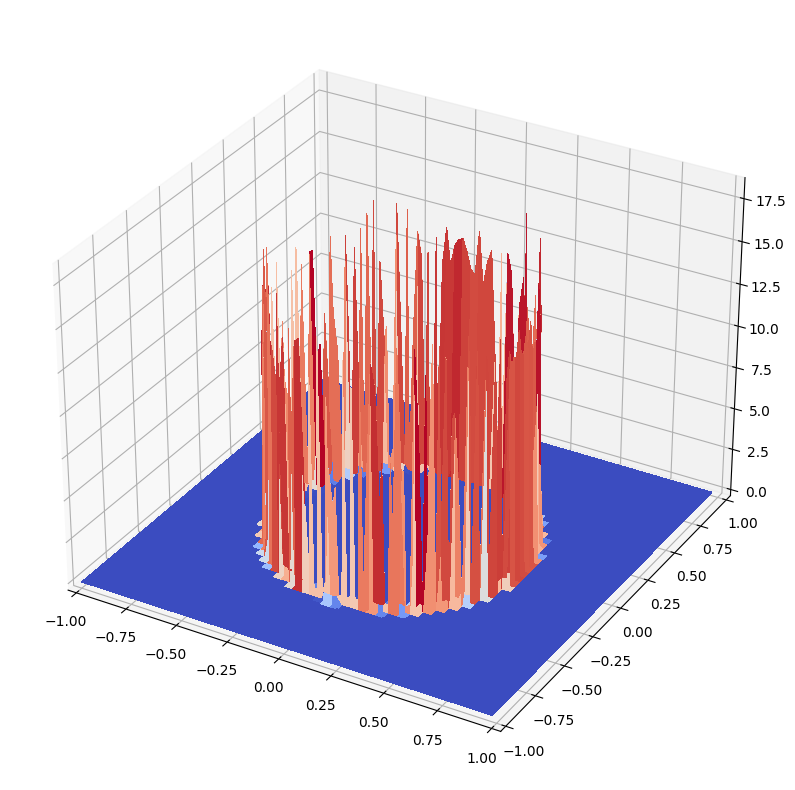}
        \includegraphics[width=0.45\linewidth]{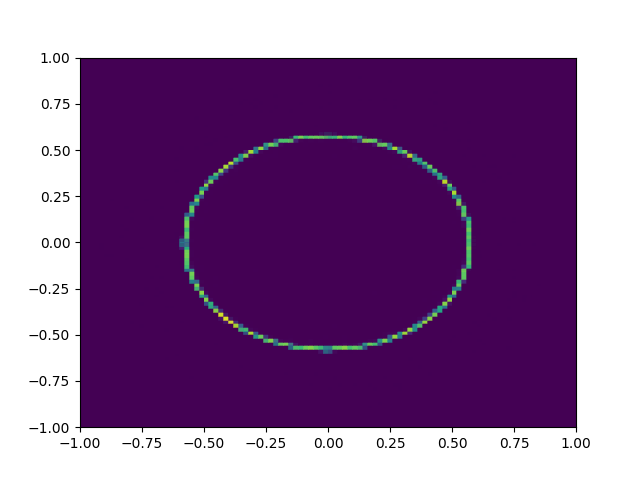}
        \caption{Forward method  $k=200, t=10$}
    \end{subfigure}
    \begin{subfigure}{0.45\textwidth}
        \includegraphics[width=0.45\linewidth]{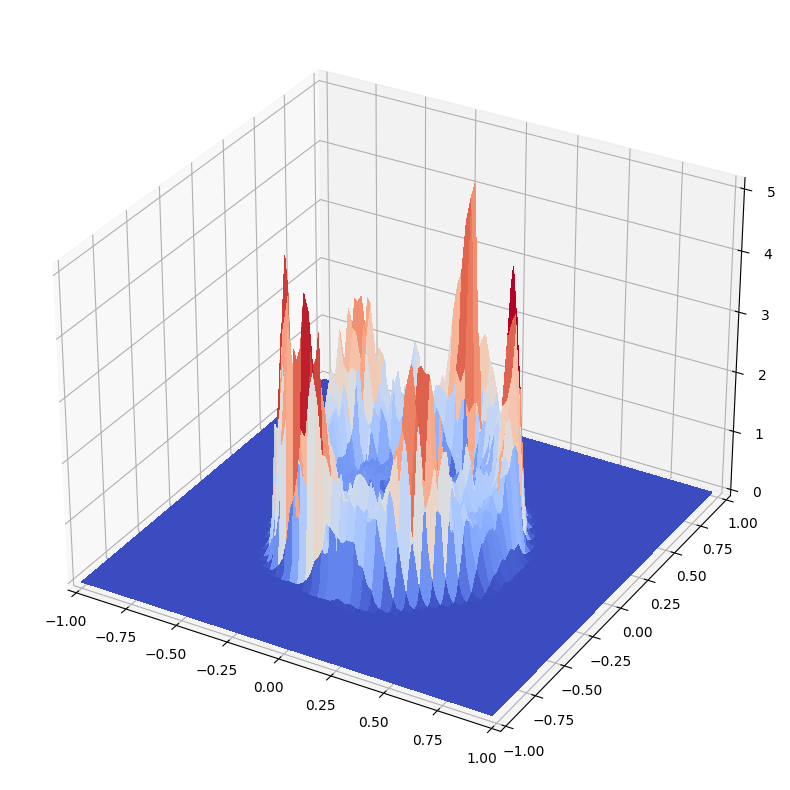}
        \includegraphics[width=0.45\linewidth]{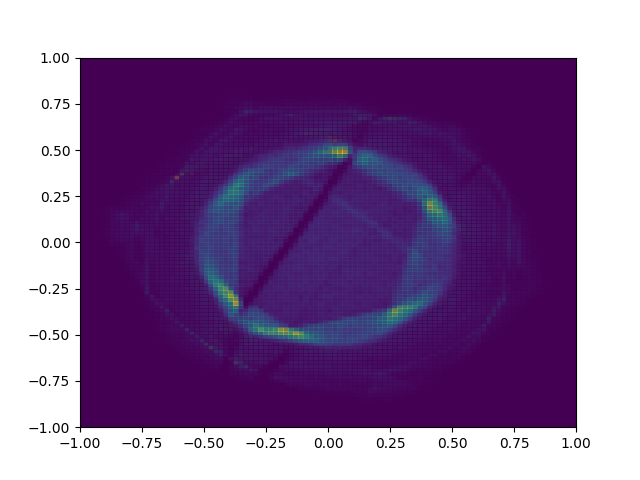}
        \caption{Backward method  $k=23, t=1.15$}
    \end{subfigure}
    \begin{subfigure}{0.45\textwidth}
        \includegraphics[width=0.45\linewidth]{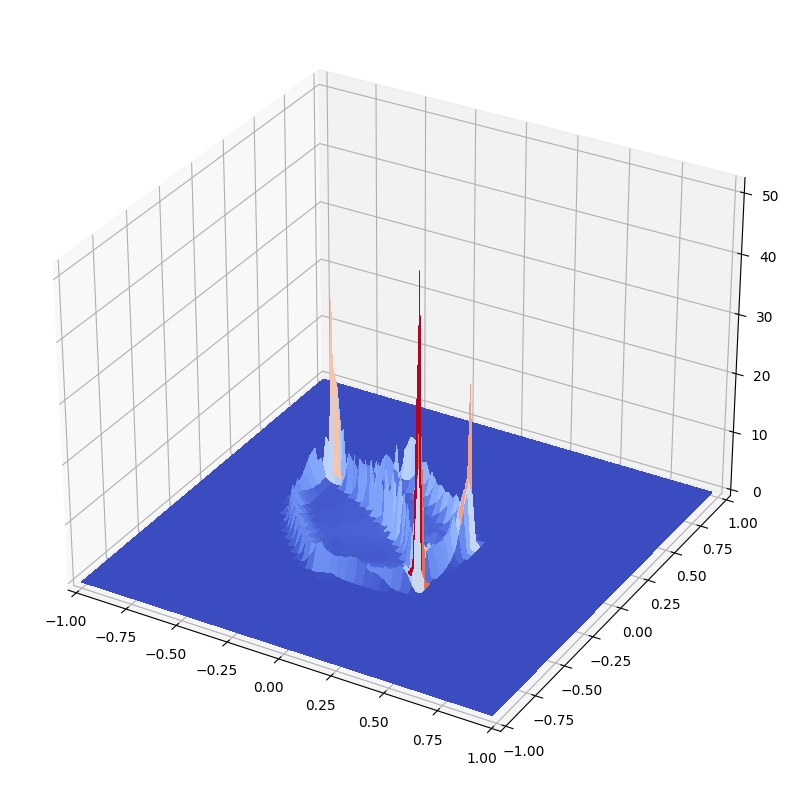}
        \includegraphics[width=0.45\linewidth]{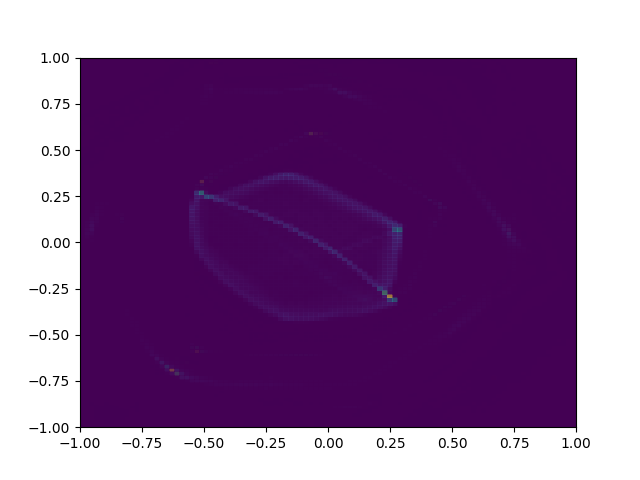}
        \caption{Backward method  $k=40, t=2$}
    \end{subfigure}
    \caption{The steady state is supported on a ring of radius 0.5.
        Backward converges faster to the steady rate but is unstable. As $k$ goes large, it cannot keep the regular ring shape and will collapse after $k>50$.}
    \label{fig:ring}
\end{figure}

\paragraph{Example 1 } We follow the setting in \citet[Section 4.3.1 ]{CarCraWanWei21}. The interaction kernel is $W(x) = \frac{\|x\|^4}{4} - \frac{\|x\|^2}{2}$, and the initial measure $P_0$ is  a Gaussian $\mN(0,0.25I)$. In this case, $\nabla_x( W*P_k ) $ becomes  $ \mE_{y\sim P_k} \left[ (\|x-y\|^2-1)(x-y) \right] $.
We use step size $a=0.05$ for both methods and show the results in Figure \ref{fig:ring}.

\paragraph{Example 2 } We follow the setting in \citet[Section 4.2.3 ]{CarCraWanWei21}. The interaction kernel is $W(x) = \frac{\|x\|^2}{2} - \ln{\|x\|}$, and the initial measure $P_0$ is $\mN(0,1)$. The unique steady state for this case is
$$P_\infty(x)= \frac{1}{\pi} \sqrt{(2-x^2)_+}.$$
The reader can refer to \citet[Section 5.3]{AlvSchMro21} for the backward method performance. As for the forward method,  $\nabla_x( W*P_k ) $ becomes  $ \mE_{y\sim P_k} \left[ x-y -\frac{1}{x-y} \right] $. Because the kernel $W$ enforces repulsion near the origin  and $P_0$ is concentrated around origin, $\nabla_x (W * P)$ will easily blow up. So the forward method is not suitable for this kind of interaction kernel.
\\

Through the above two examples, if $\nabla_x(W *P)$ is smooth, we can notice the backward method converges faster, but is not stable
when solving \eqref{eq:interaction}.
This shed light on the FB and non-FB scheme performance in Section \ref{sec:agg-diff}, \ref{sec:non-prox agg-diff}.
However, if $\nabla_x(W *P)$ has bad modality such as Example 2, the forward method loses the competitivity.

\subsection{Simulations to Aggregation–Diffusion Equation with FB scheme}\label{sec:agg-diff}
\begin{figure}[h]
    \centering
    \begin{subfigure}{0.15\textwidth}
        \centering
        \includegraphics[width=1\linewidth]{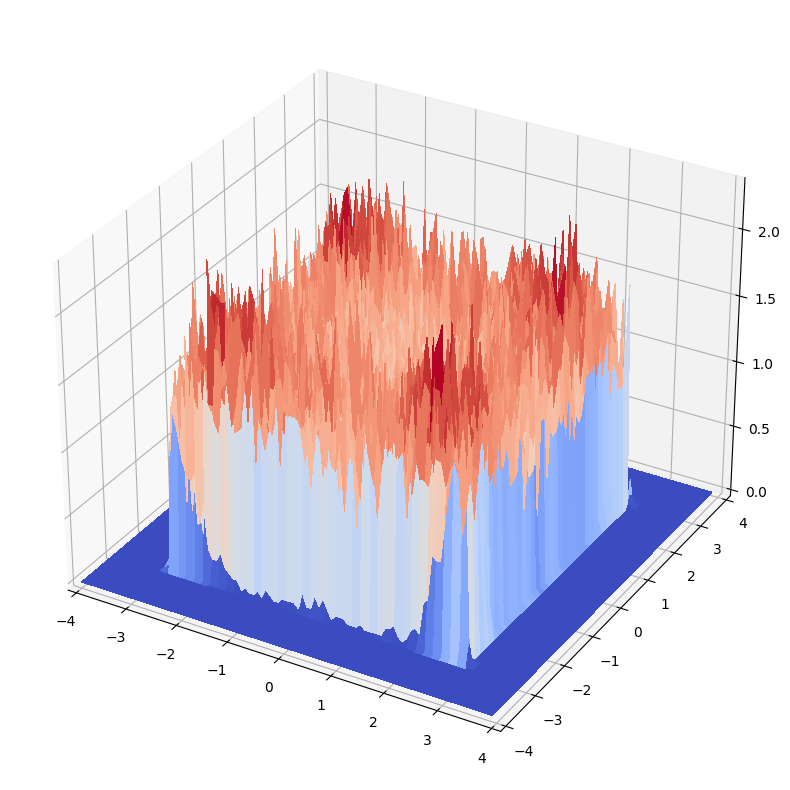}
        \caption{$k=24$}
    \end{subfigure}
    \begin{subfigure}{0.15\textwidth}
        \centering
        \includegraphics[width=1\linewidth]{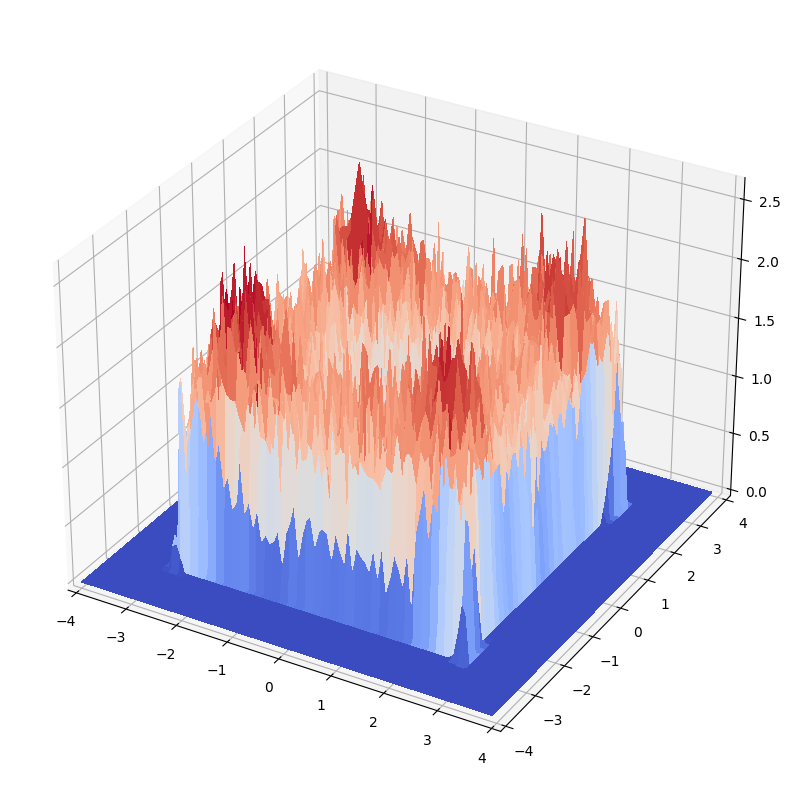}
        \caption{$k=36$}
    \end{subfigure}
    \begin{subfigure}{0.15\textwidth}
        \centering
        \includegraphics[width=1\linewidth]{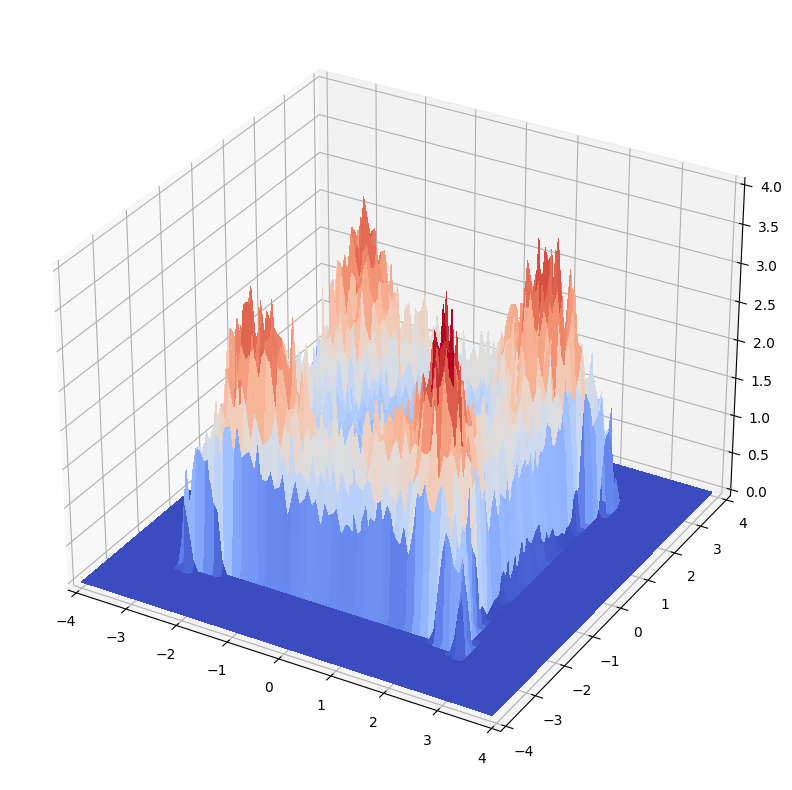}
        \caption{$k=60$}
    \end{subfigure}
    \begin{subfigure}{0.15\textwidth}
        \centering
        \includegraphics[width=1\linewidth]{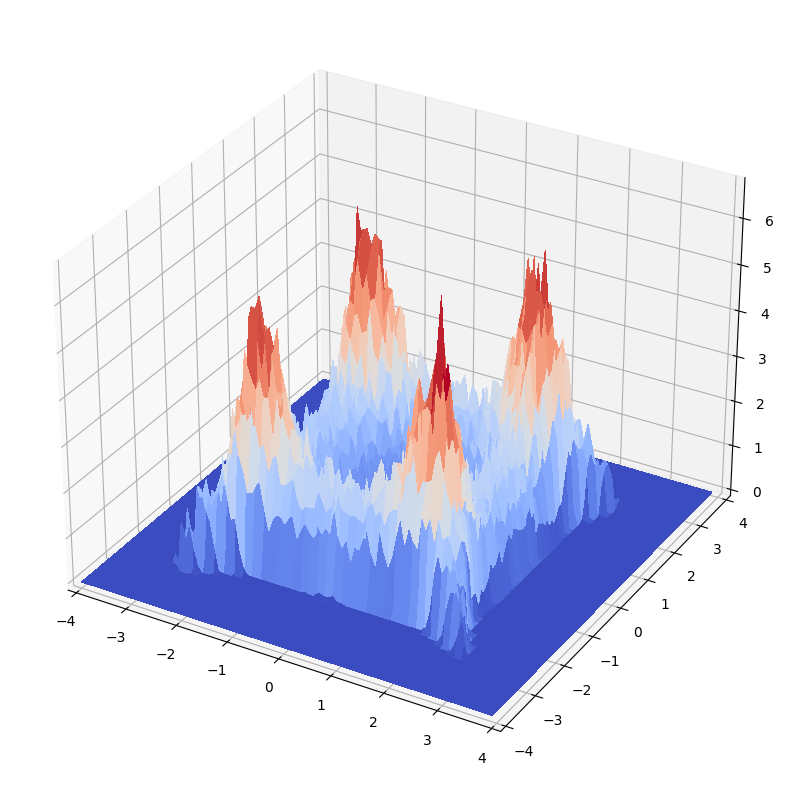}
        \caption{$k=84$}
    \end{subfigure}
    \begin{subfigure}{0.15\textwidth}
        \centering
        \includegraphics[width=1\linewidth]{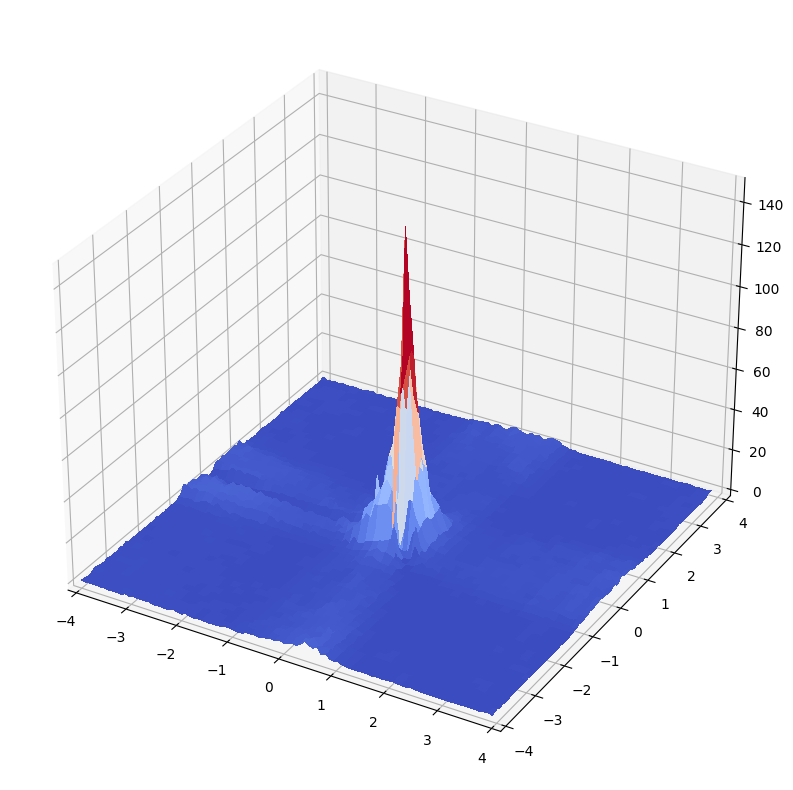}
        \caption{$k=92$}
    \end{subfigure}
    \caption{Histogram for simulated measures $P_k$ by FB scheme at different $k$.}
    \label{fig:aggreg_diffu_2d}
\end{figure}
We simulate the evolution of solutions to the following aggregation-diffusion equation:
\begin{align*}
    \partial_t P= \nabla \cdot (P \nabla W*P) + 0.1 \Delta P^m, ~ W(x)= - e^{-\|x\|^2} /\pi.
\end{align*}
This corresponds to the energy function $\mW(P) + 0.1 \mG(P)$.
There is no explicit closed-form solution for this equation except for the known singular steady state \citep{carrillo2019nonlinear}, thus we only provide qualitative results in Figure \ref{fig:aggreg_diffu_2d}.
We use the same parameters in \citet[Section 4.3.3]{CarCraWanWei21}.
The initial distribution is a uniform distribution supported on $[-3,3] \times [-3,3]$ and the JKO step size $a=0.5$. We utilize FB scheme to simulate the gradient flow for this equation with $m=3$ on $\mR^2$ space. With this choice $W(x)$, $\nabla_x( W*P_k ) $ is equal to $ \mE_{y\sim P_k} \left[ 2e^{-\|x-y\|^2} /\pi \right] $ in the gradient descent step \eqref{eq:fb1}.
And we estimate $\nabla_x( W*P_k )$ with $10^4$ samples from $P_k$.

Throughout the process, the aggregation term $\nabla \cdot (P \nabla W*P)$ and the diffusion $0.1 \Delta P^m$ adversarially exert their effects and cause the probability measure split to four pulses and converge to a single pulse in the end. Our result aligns with the simulation of discretization method \citep{CarCraWanWei21} well.

\subsection{Simulation solutions to  Aggregation-diffusion equation with non-FB scheme} \label{sec:non-prox agg-diff}
\begin{figure}[h]
    \centering
    \begin{subfigure}{0.24\textwidth}
        \centering
        \includegraphics[width=1\linewidth]{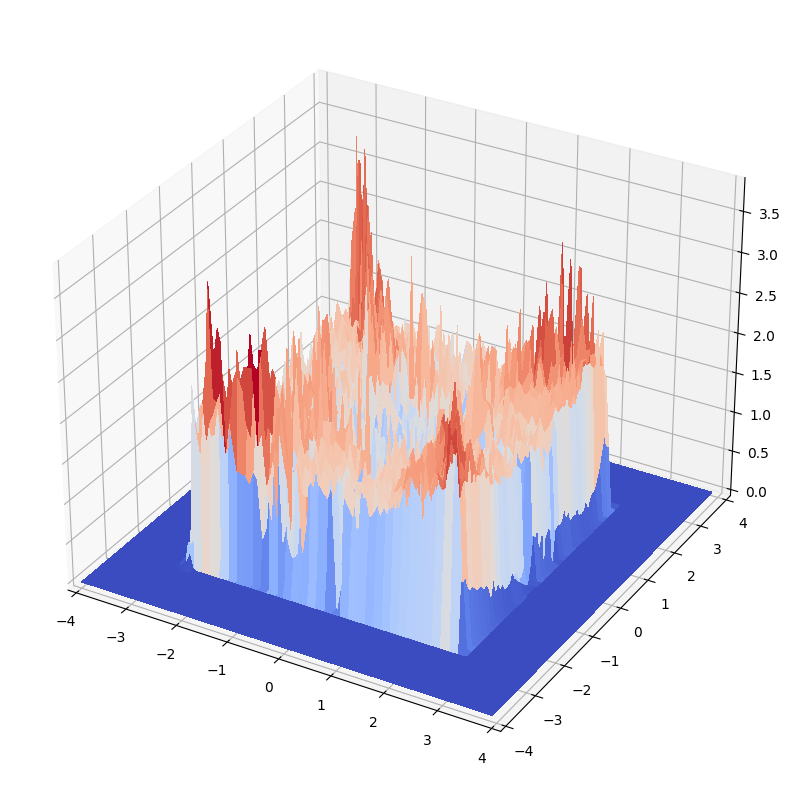}
        \caption{$k=18$}
    \end{subfigure}
    \begin{subfigure}{0.24\textwidth}
        \centering
        \includegraphics[width=1\linewidth]{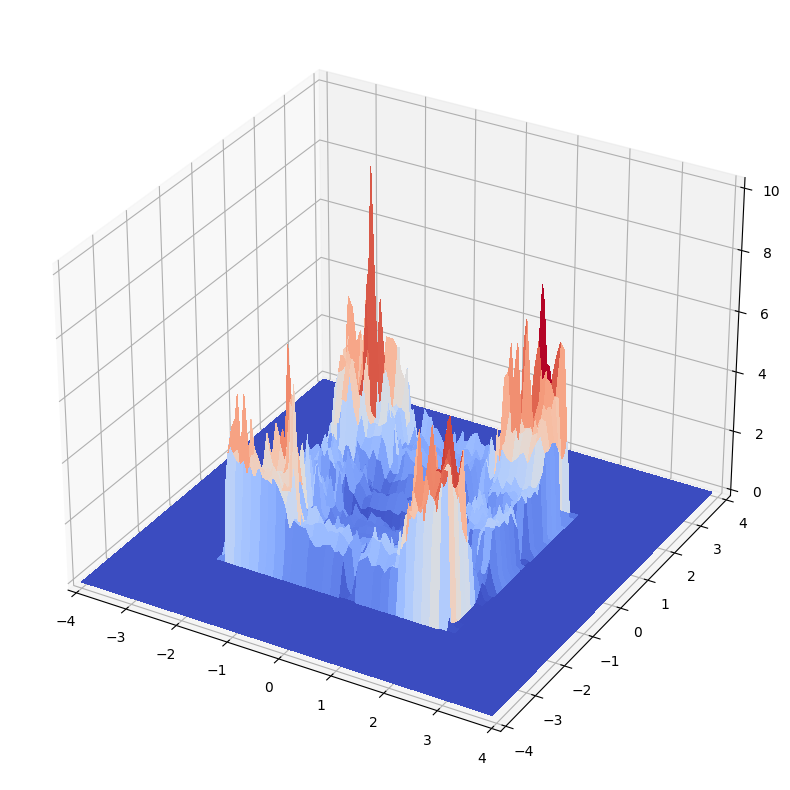}
        \caption{$k=24$}
    \end{subfigure}
    \begin{subfigure}{0.24\textwidth}
        \centering
        \includegraphics[width=1\linewidth]{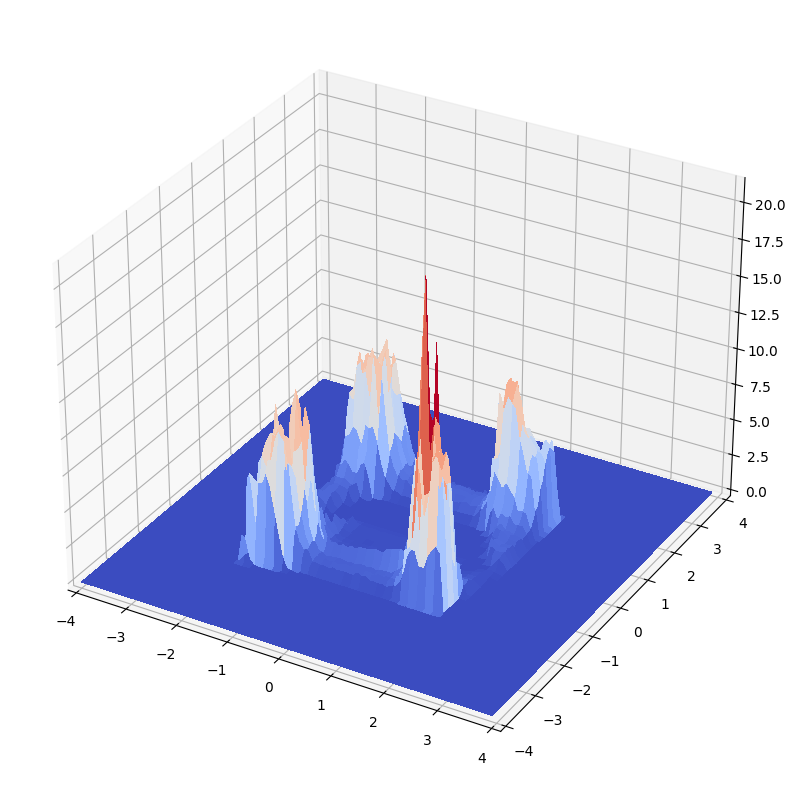}
        \caption{$k=30$}
    \end{subfigure}
    \begin{subfigure}{0.24\textwidth}
        \centering
        \includegraphics[width=1\linewidth]{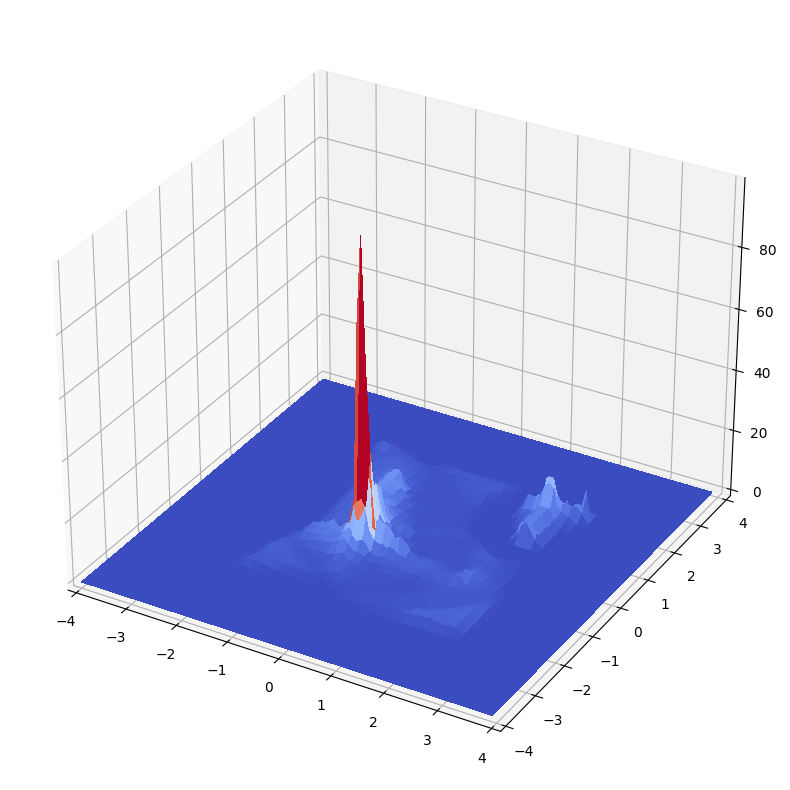}
        \caption{$k=42$}
    \end{subfigure}
    \caption{Histograms for simulated measures $P_k$ by non-FB scheme at different $k$.}
    \label{fig:aggreg_diffu_2d-nonfb}
\end{figure}

In Figure \ref{fig:aggreg_diffu_2d-nonfb}, we show the non-FB solutions to
Aggregation-diffusion equation in Section \ref{sec:agg-diff}.
FB scheme should be independent with the implementation of JKO, but in the following context, we assume FB and non-FB are both neural network based methods discussed in Section \ref{sec:method}.
Non-FB scheme reads
\begin{align*}
    P_{k+1} & =T_k\sharp P_k                                                 \\
    T_k     & = \argmin_T \left\{ \frac{1}{2a}\mathbb{E}_{P_k}[\|X-T(X)\|^2]
    +\mE_{X,Y\sim P_k} [W(T(X) - T(Y))]
    +  \mG(T,h)
    \right\},
\end{align*}
where $\mG(T,h)$ is represented by the variational formula \eqref{eq:porous_hat_h}.
We use the same step size $a=0.5$ and other PDE parameters as in Section \ref{sec:agg-diff}.

Comparing the FB scheme results in Figure \ref{fig:aggreg_diffu_2d} and the non-FB scheme results in Figure \ref{fig:aggreg_diffu_2d-nonfb}, we observe non-FB converges $1.5\times$ slower than the finite difference method \citep{CarCraWanWei21}, and FB converges  $3 \times$ slower than the finite difference method.
This may because splitting one JKO step to the forward-backward two steps removes the aggregation term effect in the JKO, and the diffusion term is too weak to make a difference in the loss.
Note at the first several $k$, both $P_k$ and $Q$ are nearly the same uniform distributions, so $h$ is nearly a constant and $T(x)$ exerts little effect in the variational formula of $\mG(P)$. Another possible reason is a single forward step for aggregation term converges slower than integrating aggregation in the backward step, as we discuss in Section \ref{sec:W only} and Figure \ref{fig:ring}.

However, FB generates more regular measures. We can tell the four pulses given by FB are more symmetric.
We speculate this is because gradient descent step in FB utilizes the geometric structure of $W(x)$ directly, but integrating $\mW(P)$ in neural network based JKO losses the geometric meaning of $W(x)$.

\section{Evaluation of the density}\label{sec:density}
In this section, we assume the solving process doesn't use forward-backward scheme, i.e. all the probability measures $P_k$ are obtained by performing JKO one by one. Otherwise, the map $I- a \nabla_x( W*P_k )=I-\mE_{y \sim P_k}\nabla_x( W(x-y) )$ includes an expectation term and becomes intractable to push-backward particles to compute density.

If $T$ is invertible, these exists a standard approach, 
{which we present here for completeness,} to evaluate the density of $P_k$ \citep{AlvSchMro21, MokKorLiBur21} through the change of variables formula. More specifically, we assume $T$ is parameterized by the gradient of an ICNN $\varphi$ that is assumed to be strictly convex. Thus we can guarantee that the gradient $\nabla \varphi$ invertible. To evaluate the density $P_k(x_k)$ at point $x_k$, we back propagate through the sequence of maps $T_k=\nabla \varphi_k, \ldots, T_1=\nabla \varphi_1$ to get
\[
    x_i= T_{i+1}^{-1}\circ T_{i+2}^{-1}\circ \cdots \circ T_k^{-1}(x_k).
\]
The inverse map $T_j^{-1} = (\nabla \varphi_j)^{-1} = \nabla \varphi_j^*$ can be obtained by solving the convex optimization
\begin{align}\label{eq:particle_history}
    x_{j-1}= \argmax_{x \in \mR^n} \langle x , x_j\rangle - \varphi _j (x).
\end{align}

Then, by the change of variables formula, we obtain
\begin{align} \label{eq:density}
    \log [dP_k(x_k)] = \log [dP_0 (x_0)]- \sum_{i=1}^k \log
    \left|\nabla^2 \varphi_i (x_{i-1}) \right|,
\end{align}
where $\nabla^2 \varphi_i(x_{i-1})$ is the Hessian of $\varphi_i$ and $|\nabla^2 \varphi_i (x_{i-1})|$
is its determinant. By iteratively solving \eqref{eq:particle_history} and plugging the resulting $x_j$ into \eqref{eq:density}, we can recover the density $dP_k(x_k)$ at any point.

\section{Additional experiment results and discussions}\label{sec:addition}

\subsection{Computational time}\label{sec:time}
The forward step \eqref{eq:fb1} takes about 14 seconds to pushforward one million points.

Other than learning generative model, assume each JKO step involves 500 iterations, the number of iterations $J_2=3$, $J_3=2$, then the training of each JKO step \eqref{eq:fb2} takes around $15$ seconds.

For learning image generative model, assume  $J_2=1$, $J_3=5$, then the training of each JKO step \eqref{eq:fb2} takes around 20 minutes.

\subsection{Learning of function $h$}
The learning of the function  $h$ is crucial because it determines the effectiveness of variational formula. 
In our KL divergence and generalized entropy variational formulas, the optimal $h$ is equal to $T\sharp P_k /\Gamma$, which can have large Lipschitz constant in some high dimensional applications and become difficult to approximate. To tackle this issue, we replace $h$ by $\exp(\bar h)-1$, thus the optimal $\bar h$ is $\log(h+1)$, whose Lipschitz constant is much weakened. We apply this trick in Section \ref{sec:sampling} and observe the improved performance.

In image tasks, $h$ works like a discriminator in GAN. A typical problem in GAN is that the discriminator can be too strong to let generator keep learning. To avoid this, we add the spectral normalization in $h$ such that the Lipschitz of $h$ is bounded by 1.

\subsection{Convergence comparison with the same number of JKO steps}\label{sec:gmm_same_step}
\begin{figure}[h]
    \centering
    \includegraphics[width=0.3\linewidth]{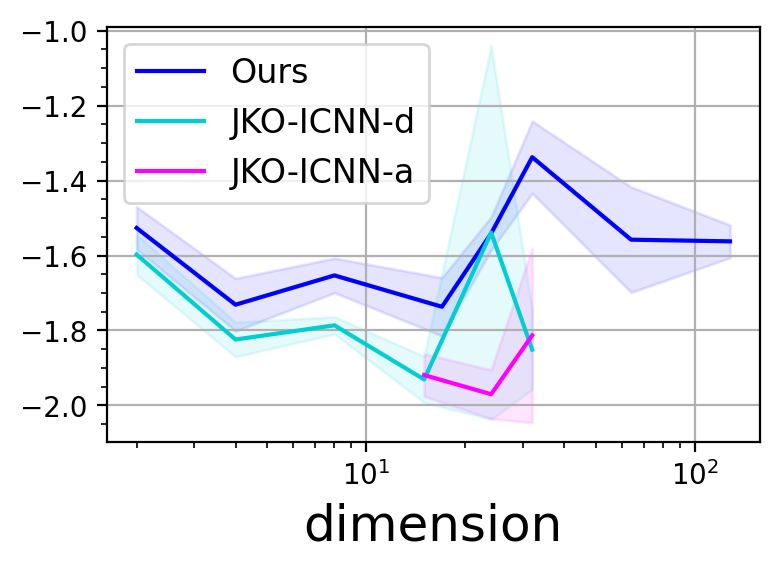}
    \caption{Quantitative comparison in converging to GMM with the constraint of performing 40 JKO steps for all methods. We calculate the kernelized Stein divergence between the generated distribution and the target distribution.
    }
    \label{fig:gmm_ksd_full_run}
\end{figure}

In this section, we show the convergence comparison under the constraint of performing same number of JKO steps for all methods.
The result is in Figure \ref{fig:gmm_ksd_full_run}.
We repeat the experiment for 5 times with the same global random seed $1,2,3,4,5$ for all methods. JKO-ICNN shows large variance and instability after longer run in high dimension.
Specifically, we observe that at random seed 2 in dimension 24, JKO-ICNN-d converges for the first 19 JKO steps and then suddenly diverges, causing the occurrence of an extreme point. The similar instability issue is also reported in \citet[Figure 3]{bonet2021sliced}. With the same random seeds, through 40 JKO steps, we don't observe this instability issue using our method. 

\section{Experiments implementation details other than image} \label{sec:params}
 Our experiments are conducted on GeForce RTX 3090 or RTX A6000. We always make sure the comparison is conducted on the same GPU card when comparing training time with other methods.
Our code is written in Pytorch-Lightning \citep{falcon2020framework}. We use other wonderful python libraries including W\&B~\citep{wandb}, hydra \citep{Yadan2019Hydra}, seaborn \citep{Waskom2021}, etc. We also adopt the code given by \citet{MokKorLiBur21} for some experiments. For fast approximation of $\log \det \nabla^2 \varphi$, we adapt the code given by \citet{huang2020convex} with default parameters therein.

Without further specification, we use the following parameters:
\begin{itemize}
    \item  The number of iterations: $J_1=600$. $J_2=3$. $J_3=1$.
    \item The batch size is fixed to be $M=100$.
    \item The learning rate is fixed to be $0.001$.
    \item All the activation functions are set to be PReLu.
    \item $h$ has 3 layers and 16 neurons in each layer.
    \item $T$ has 4 layers and 16 neurons in each layer.
\end{itemize}

The transport map $T$ can be parametrized in different ways. We use a residual MLP network for it in Section \ref{sec:sampling}, \ref{sec:ou}, \ref{sec:bayesian}, \ref{sec:agg-diff}, \ref{sec:W only}, and the gradient of a strongly convex ICNN in Section \ref{sec:porous},
\ref{sec:non-prox agg-diff}. 
Except image task, the dual test function $h$ is always a MLP network with quadratic or sigmoid actication function in the final layer to promise $h$ is positive. \\
The networks $T$ and $h$ in Section \ref{sec:image} are chosen to be UNet and a normal CNN.

\subsection{Calculation of error criteria}
\paragraph{Sampling from GMM} 
We estimate the kernelized Stein discrepancy (KSD) following the author's instructions \citep{liu2016kernelized}. We draw $N$ samples $X_1,\ldots,X_N$ from each method, and estimate KSD as 
\begin{align*}
    \text{KSD}(P,Q) =  \frac{1}{N(N-1)} \sum_{1\leq i \neq j \leq N} u_Q(X_i,X_j),
\end{align*}
where 
\begin{align*}
 u_Q(x,x') & = s_{q}(x)^{\top} k\left(x, x^{\prime}\right) s_{q}\left(x^{\prime}\right)+s_{q}(x)^{\top} \nabla_{x^{\prime}} k\left(x, x^{\prime}\right)+ \nabla_{x} k\left(x, x^{\prime}\right)^{\top} s_{q}\left(x^{\prime}\right)+\operatorname{trace}\left(\nabla_{x, x^{\prime}} k\left(x, x^{\prime}\right)\right),\\
 s_Q &= \nabla_x \log Q(x) =\frac{\nabla_x Q(x)}{ Q(x)}. 
\end{align*}

We choose the kernel $\phi$ to be the RBF kernel and use the same bandwidth for all methods.  We fix $N=1\times 10^{5}$,

\paragraph{OU process} 
For each method, we  draw $5\cdot 10^5$ samples from ${P}_t$ and calculate the empirical mean $\widetilde{\mu}_t$ and covariance $\widetilde{\Sigma}_t$. Then we calculate the SymKL between $\mN( \tilde{\mu}_t, \tilde{\Sigma}_t)$ and the exact solution. 



\paragraph{Porous media equation} 
We calculate the density of $P_k$ according to Section \ref{sec:density} and estimate the SymKL using Monte Carlo according to the instructions in \citet{MokKorLiBur21}.



\subsection{Sampling from Gaussian Mixture Models (Section \ref{sec:sampling}
    )}

\paragraph{Two moons}
We run $K=10$ JKO steps with $J_2=6, J_3=1$ inner iterations. $h$ has 5 layers. $T$ has 4 layers.
\paragraph{GMM}
The mean of Gaussian components are randomly sampled from $\text{Uniform}([-\ell/2,\ell/2]^n)$. $J_3=2$. The map $T$ has dropout in each layer with probability 0.04.
The learning rate of our method is $1\cdot 10^{-3}$ for the first 20 JKO steps and $4\cdot 10^{-4}$ for the last 20 JKO steps. The learning rate of JKO-ICNN is $5\cdot 10^{-3}$ for the first 20 JKO steps, and then $2\cdot 10^{-3}$ for the rest steps. The batch size is 512 and each JKO step runs 1000 iterations for all methods. The rest parameters are in Table \ref{table:gmm_param}.

\begin{table}[]
    \caption{Hyper-parameters in the GMM convergence experiments.
    }
    \begin{center}
        \begin{small}
            \begin{tabular}{cccccccc}
                \toprule
                                                &                               & \multicolumn{4}{c}{ { Our methods }} & \multicolumn{2}{c}{ { JKO-ICNN }}                                         \\
                \multirow{-2}{*}{ {Dimension }} & \multirow{-2}{*}{ {$\ell$  }} & $T$ width                            & $T$ depth                         & $h$ width & $h$ depth & width & depth \\
                \toprule
                2                               & 5                             & 8                                    & 3                                 & 8         & 3         & 256   & 2     \\
                4                               & 5                             & 32                                   & 4                                 & 32        & 3         & 384   & 2     \\
                8                               & 5                             & 32                                   & 4                                 & 32        & 4         & 512   & 2     \\
                15                              & 3                             & 64                                   & 4                                 & 64        & 4         & 1024  & 2     \\
                17                              & 3                             & 64                                   & 4                                 & 64        & 4         & 1024  & 2     \\
                24                              & 3                             & 64                                   & 5                                 & 64        & 4         & 1024  & 2     \\
                32                              & 3                             & 64                                   & 5                                 & 64        & 4         & 1024  & 2     \\
                64 & 2 & 128 & 5 & 128 & 4 & - & -\\
                128 & 1.5 & 128 & 5 & 128 & 4 & - & - \\
                \bottomrule
            \end{tabular}
        \end{small}
    \end{center}
    \vskip -0.1in
    \label{table:gmm_param}
\end{table}

\subsection{Ornstein-Uhlenbeck Process (Section \ref{sec:ou})}
{We use nearly all the same hyper-parameters as \citet{MokKorLiBur21}, including learning rate, hidden layer width, and the number of iterations per JKO step. Specifically,} we use a residual feed-forward NN to work as $T$, i.e. without activation function.
$h$ and $T$ both have 2 layers and 64 hidden neurons per layer for all dimensions.
We also train them for $J_1=500$ iterations per each JKO with learning rate $0.005$. The batch size is $M=1000$.
\subsection{Bayesian Logistic Regression (Section \ref{sec:bayesian})}
Same as \citet{MokKorLiBur21}, we use JKO step size $a=0.1$ and calculate the log-likelihood and accuracy with 4096 random parameter samples. The rest parameters are in Table \ref{table:bayesian_param}.
\begin{table*}[]
    \centering
    \caption{Bayesian logistic regression accuracy and log-likelihood full results.} \label{tab: full bayesian}
    \begin{tabular}{ccccccccc}
        \hline                        &                                   &                                    & \multicolumn{3}{c}{ { Accuracy }} & \multicolumn{3}{c}{ { Log-Likelihood }}                                                                    \\
        \multirow{-2}{*}{ {Dataset }} & \multirow{-2}{*}{ {\# features }} & \multirow{-2}{*}{ {dataset size }} & Ours                              & \text{JKO-ICNN-d}                       & \text { SVGD } & Ours     & \text { JKO-ICNN-d} & \text { SVGD } \\
        \hline \text { covtype }      & 54                                & 581012                             & {0.753}                           & 0.75                                    & 0.75           & -0.528   & {-0.515}            & {-0.515}       \\
        \text { splice }              & 60                                & 2991                               & 0.84                              & 0.845                                   & {0.85}         & -0.38    & -0.36               & {-0.355}       \\
        \text { waveform }            & 21                                & 5000                               & {0.785}                           & 0.78                                    & 0.765          & {-0.455} & -0.485              & {-0.465}       \\
        \text { twonorm }             & 20                                & 7400                               & {0.982}                           & 0.98                                    & 0.98           & {-0.056} & -0.059              & -0.062         \\
        \text { ringnorm }            & 20                                & 7400                               & 0.73                              & {0.74}                                  & {0.74}         & -0.5     & -0.5                & -0.5           \\
        \text { german }              & 20                                & 1000                               & {0.67}                            & {0.67}                                  & 0.65           & {-0.59}  & -0.6                & -0.6           \\
        \text { image }               & 18                                & 2086                               & {0.866}                           & 0.82                                    & 0.815          & {-0.394} & -0.43               & -0.44          \\
        \text { diabetis }            & 8                                 & 768                                & {0.786}                           & 0.775                                   & 0.78           & {-0.45}  & {-0.45}             & -0.46          \\
        \text { banana }              & 2                                 & 5300                               & {0.55}                            & {0.55}                                  & 0.54           & -0.69    & -0.69               & -0.69          \\
        \hline
    \end{tabular}
\end{table*}
\begin{table}[]
    \caption{Hyper-parameters in the Bayesian logistic regression.
    }
    \begin{center}
        \begin{small}
            \begin{tabular}{cccccccccc}
                \toprule
                Dataset  & $K$ & $M$  & $J_1$ & $T$ width & $T$ depth & $h$ width & $h$ depth & $T$ learning rate & $h$ learning rate \\
                \toprule
                covtype  & 7   & 1024 & 7000  & 128       & 4         & 128       & 3         & $2\cdot 10^{-5}$  & $2\cdot 10^{-5}$  \\
                splice   & 50  & 1024 & 400   & 128       & 5         & 128       & 4         & $1\cdot 10^{-4}$  & $1\cdot 10^{-4}$  \\
                waveform & 5   & 1024 & 1000  & 32        & 4         & 32        & 4         & $1\cdot 10^{-5}$  & $5\cdot 10^{-5}$  \\
                twonorm  & 15  & 512  & 800   & 32        & 4         & 32        & 3         & $1\cdot 10^{-3}$  & $1\cdot 10^{-3}$  \\
                ringnorm & 9   & 1024 & 500   & 32        & 4         & 32        & 4         & $1\cdot 10^{-5}$  & $1\cdot 10^{-5}$  \\
                german   & 14  & 800  & 640   & 32        & 4         & 32        & 4         & $2\cdot 10^{-4}$  & $2\cdot 10^{-4}$  \\
                image    & 12  & 512  & 1000  & 32        & 4         & 32        & 4         & $1\cdot 10^{-4}$  & $1\cdot 10^{-4}$  \\
                diabetis & 16  & 614  & 835   & 32        & 4         & 32        & 3         & $1\cdot 10^{-4}$  & $1\cdot 10^{-4}$  \\
                banana   & 16  & 512  & 1000  & 16        & 2         & 16        & 2         & $5\cdot 10^{-4}$  & $5\cdot 10^{-4}$  \\
                \bottomrule
            \end{tabular}
        \end{small}
    \end{center}
    \vskip -0.1in
    \label{table:bayesian_param}
\end{table}
\subsection{Porous media equation (Section \ref{sec:porous})} \label{sec:disc_cvxopt}
We use rejection sampling \citep{eckhardt1987monte} to sample from $P_0$ because its computational time is more promising than MCMC methods. However, the rejection sampling acceptance rate is expected to be exponentially small \citep[Ch 29.3]{mackay2003information} in dimension, and empirically it's intractable when $n > 6$. So we only give the results for $n\leq 6$.

In the experiment, $h$ have 4 layers and 16 neurons in each layer with CELU activation functions except the last layer, which is activated by PReLU. To parameterize the map, we adopt DenseICNN  \citep{KorEgiAsaBur19} structure with width 64, depth 2 and rank 1. The batch size is $M=1024$. Each JKO step runs $J_1=1000$ iterations. The learning rate for both $\varphi$ and $h$ is $1\cdot 10^{-3}$. $J_3=1$ for dimension 3 and $J_3=2$ for dimension 6.

\subsection{Aggregation-diffusion equation (Section \ref{sec:agg-diff} and \ref{sec:non-prox agg-diff})}

Each JKO step contains $J_1=200$ iterations.
The batch size is $M=1000$.

\section{Image experiment details}
\label{app:img-exp}

\subsection{Hyperparameters and network architecture}
We use Adam optimizer with learning rate $2\times 10^{-4}$ and other default settings in PyTorch library. We choose $J_2=1, J_3=5.$
Our $h$ network follows the architecture of ResNet classifier network~\citep{he2016deep}. More specially, our module uses two downsampling modules, which results in three feature map resolution~($32\times 32, 16\times 16, 8\times 8$). We use two convolutional residual blocks for each resolution and pass the features extracted from at $8\times 8$ resolution into a 2-layer MLP. We use $128$ channels for CNN and $128$ hidden neurons for the MLP. Similar to training generative adversarial networks, we found adding regularizers on $h$ network can help stabilize training. Thus, we apply the spectral normalization~\citep{miyato2018spectral} on $h$ network.

Our framework requires the $T_{k}$ networks to approximate mappings between same dimensional data spaces. 
Our network architecture follows the backbone of PixelCNN++~\citep{salimans2017pixelcnn++}, which can be viewed as a modified U-Net~\citep{ronneberger2015u} based on Wide ResNet~\citep{zagoruyko2016wide}. 
More specifically, we use 3 downsampling and 3 upsampling modules, which results in four feature map resolutions ($32\times32, 16\times16,8\times8, 4\times4$). 
At each resolution, we have two convolutional residual blocks. 
We use $64, 128, 256, 512$ channels for as image resolution decreases. 

\begin{figure}
    \centering
    \includegraphics[width=0.3\linewidth]{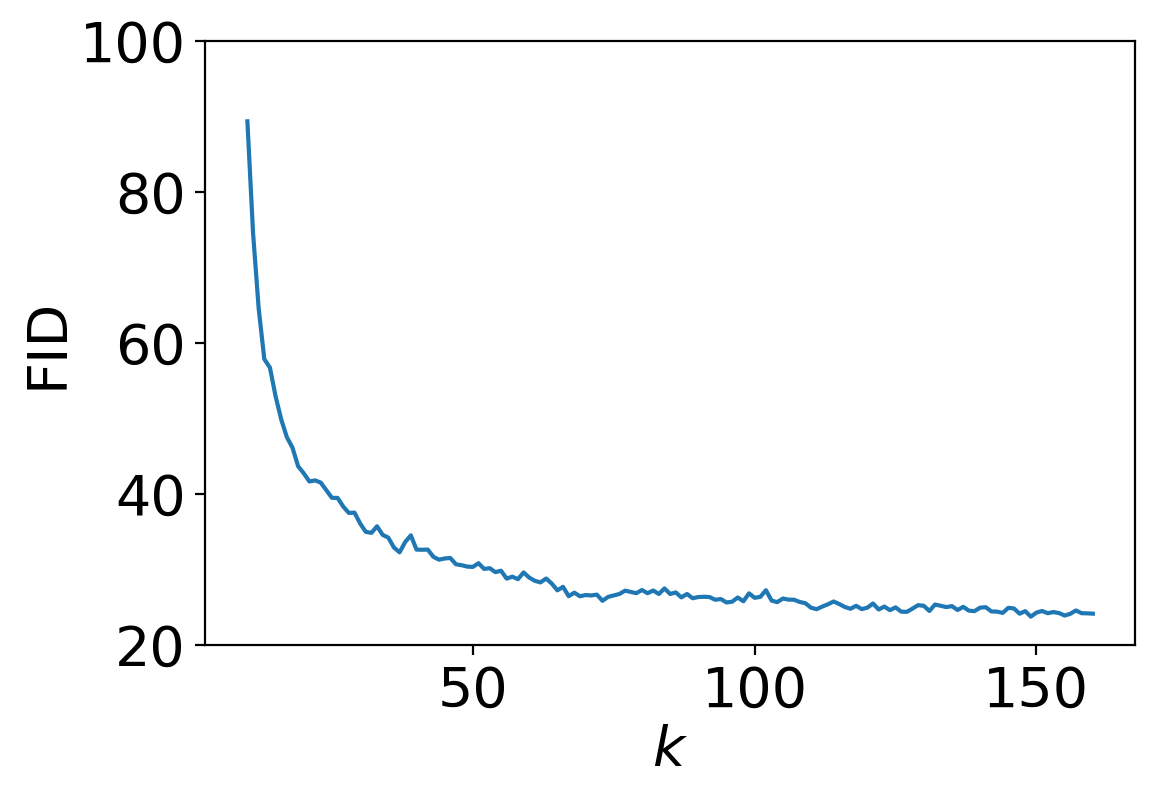}
    \caption{The FID score  converges as $k$ increases
    on CIFAR10 datset.
    }
    \label{fig:app-cifar-fid}
\end{figure}

Here are more training details:
\begin{itemize}
    \item We resize MNIST image to $32\times 32$ resolution so that we $h, T_{k}$ networks can work on both MNIST and CIFAR10 with small modification of input channel.
    \item We use random horizontal flips during training for CIFAR10.
    \item We use batch size $M=128$.
    \item On CIFAR10, we use implementation from \textit{torch-fidelity}\footnote{\url{https://github.com/toshas/torch-fidelity}} to calculate FID scores with $50$k samples.
    \item The JKO step size $a$ controls the divergence between $P_{k}$ and $P_{k+1}$. We observe training with large  $a$ has unstable issues and mode collapse, a small $a$ suffers from slower convergence. We found $a = 5$ works well on both MNIST and CIFAR10 datasets. 
    \item We use 10 epochs to train each $P_k$, we notice $P_{30}$ generates realistic images when $a=5.0$. However, we find FID score decreases as $k$ increases. 
    We present the change of FID score of samples from different $P_k$ 
    in Figure~\ref{fig:app-cifar-fid}.
\end{itemize}

\begin{figure}
    \centering
    \includegraphics[width=0.3\linewidth]{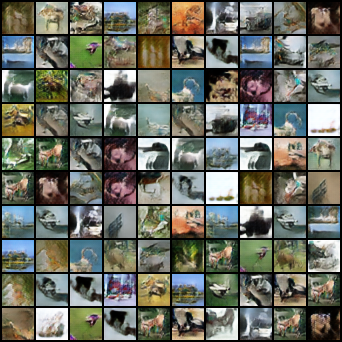}
    \caption{Mode collapsing in GANs.}
    \label{fig:gan-mode-col}
\end{figure}

\subsection{More Comparison}

\textbf{Comparison with GANs.} As we use Jensen-Shannon divergence in our scheme, JKO-Flow specializes to Jensen-Shannon GAN when $a \rightarrow \infty, K=1$. However, we found training with $a \rightarrow \infty, K=1$ is very unstable and suffer mode collapsing occasionally. Though training GANs can not recover the gradient flow from noise to image, it is interesting to compare JKO-Flow and GANs in term of sampling quality. To make a fair comparison, we instantiate generator network as the same as $T_{k}$ network and discriminator as $h$ for GANs. We note such choice is not optimal for GAN since generators in existing works usually map a lower dimensional Gaussian noise into images instead of mapping 
from
same dimensional space. We believe the comparison and JKO-Flow scheme may help future research when modeling mapping between same dimensional data spaces. 
As shown in Table~\ref{table:fid-compare}, JKO-Flow enjoys better sample qualities. 
Empirically we found training GANs is more challenging when latent space is relative large and with more complex generator networks as mode collapsing becomes more common.
We find the additional Wasserstein distance loss in JKO-Flow can be viewed a regularizer to avoid mode collapsing because $T_k$ will receive large penalty if it maps all inputs into a local minimal.
However, one shortcoming of our method is the scheme of JKO-Flow needs to model a sequence of generators instead of one generate that push $P_0$ particles into $Q$, and small step size controlled by $a$ resulted in slower convergence and more training time.

\begin{table}[]
    \begin{center}
        \begin{small}
            \begin{tabular}{c|c}
                \toprule
              Method  & FID score $\downarrow$ \\
                \toprule
                GAN~(JKO-Flow with $a \rightarrow \infty, K=1$) & $\geq 80$ \\ 
                WGAN-GP & $62.3$\\
                SN-GAN & $43.2$\\
                JKO-Flow~
                & \textbf{23.1} \\
                \bottomrule
            \end{tabular}
        \end{small}
    \end{center}
    \caption{Comparison between JKO-Flow and various GANs. 
    The generator and discriminator networks in GANs follow same architecture of $P_k$ and $h$ network in JKO-Flow.}
    \label{table:fid-compare}
\end{table}

\textbf{Comparison with more generative models based on gradient flows and optimal transport maps}. Most existing works in this line focus on the latent spaces of pre-trained autoencoders~\citep{seguy2017large, an2019ae,an2020ae,MakTagOhLee20, KorEgiAsaBur19}. The approach reduces burden of training gradients and optimal transport maps since tasks of modeling complex image modality and interactions between pixels are left to pre-trained decoders partially. We note the recent work~\citet{rout2021generative} investigates mappings between distributions located on the spaces with same dimensionality or unequal dimensionality. However, they only demonstrate the unconditional image generative model based on an embedding from a lower dimensional Gaussian distribution to image distributions. In contrast, we show JKO-Flow can learn complex mappings between both high dimensional distribution and achieve encouraging performance when applying such learned mappings in the challenging image generation task without additional conditional signal. We include more comparison in Table \ref{table:more-comp}.

\begin{table}[]
    \begin{center}
        \begin{small}
            \begin{tabular}{c|c|c}
                \toprule
              Method  & FID score $\downarrow$ & Inception Score $\uparrow$\\
                \toprule
                AE-OT~\citep{an2019ae} & 28.5 & - \\
                AE-OT-GAN~\citep{an2020ae} & 17.1 & - \\
                OTM~\cite{rout2021generative} & 20.69 & 7.41 $\pm$ 0.11 \\
                JKO-Flow~ & 23.1 & 7.48 $\pm$ 0.12 \\
                \bottomrule
            \end{tabular}
        \end{small}
    \end{center}
    \caption{More comparison among generative models on CIFAR10.}
    \label{table:more-comp}
\end{table}

\subsection{More generated samples and trajectories}

\begin{figure}
    \centering
    \includegraphics{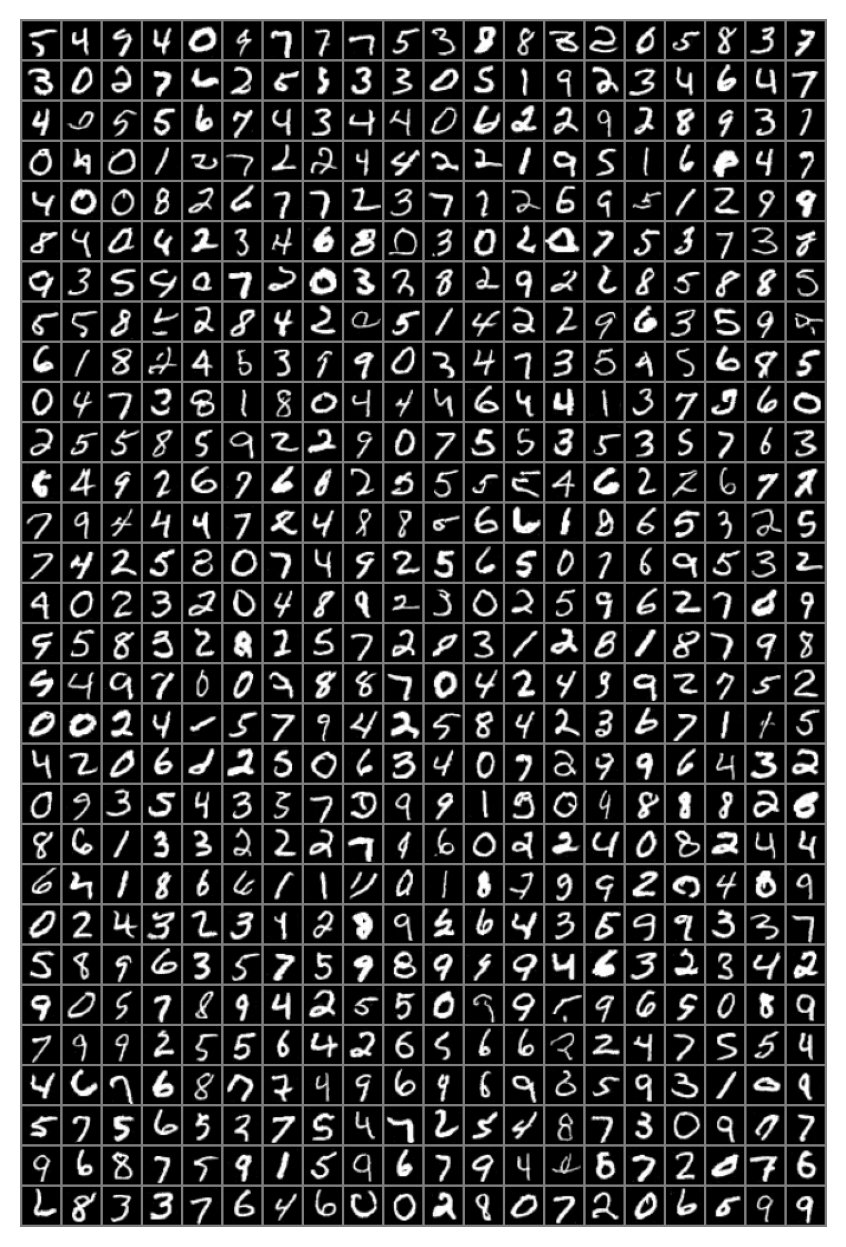}
    \caption{More MNIST sample from JKO-Flow}
    \label{fig:app-mnist}
\end{figure}

\begin{figure}
    \centering
    \includegraphics{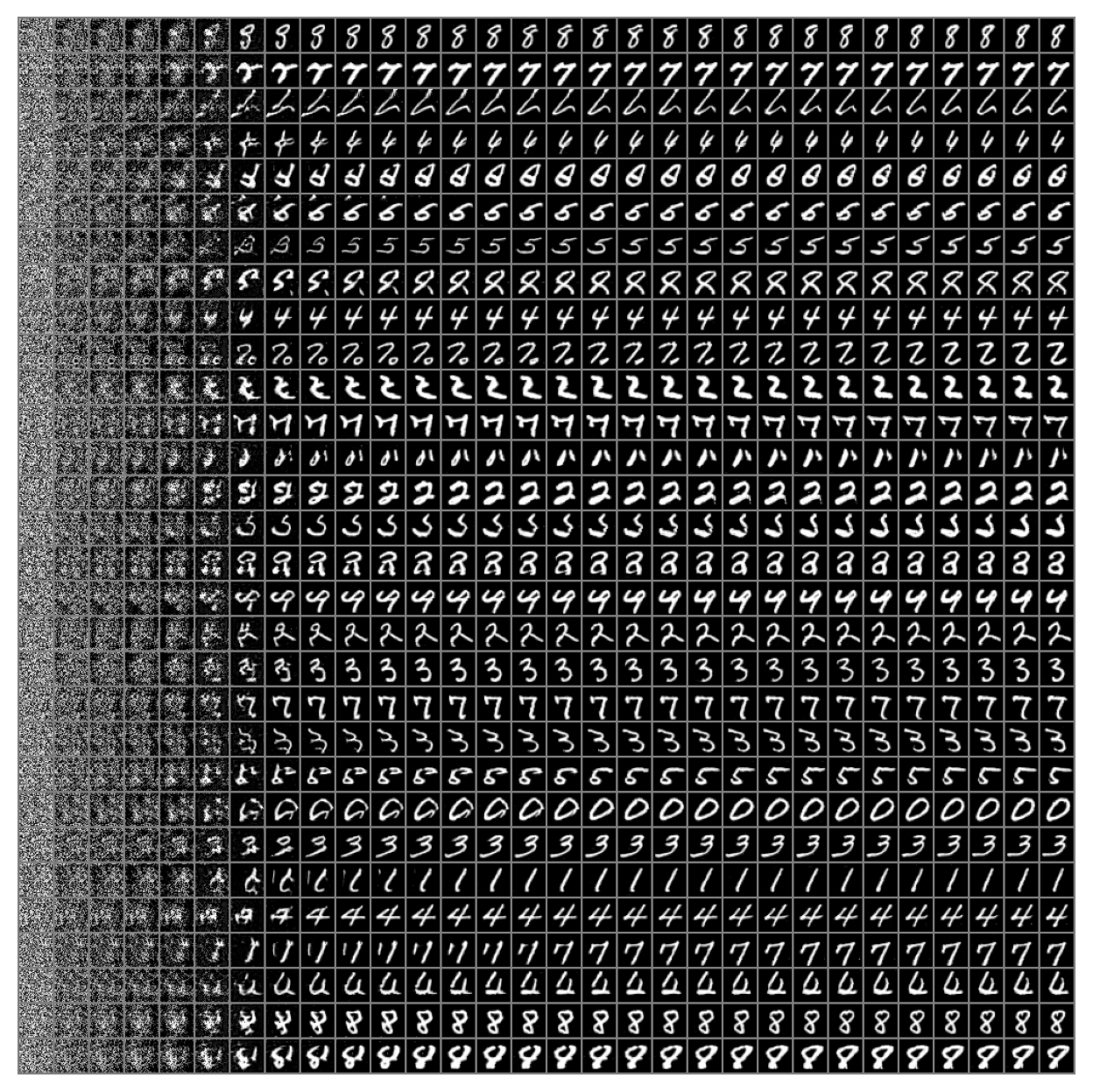}
    \caption{More MNIST trajectories from JKO-Flow with $K=1$ to $K=30$.}
    \label{fig:app-mnist-traj}
\end{figure}

\begin{figure}
    \centering
    \includegraphics{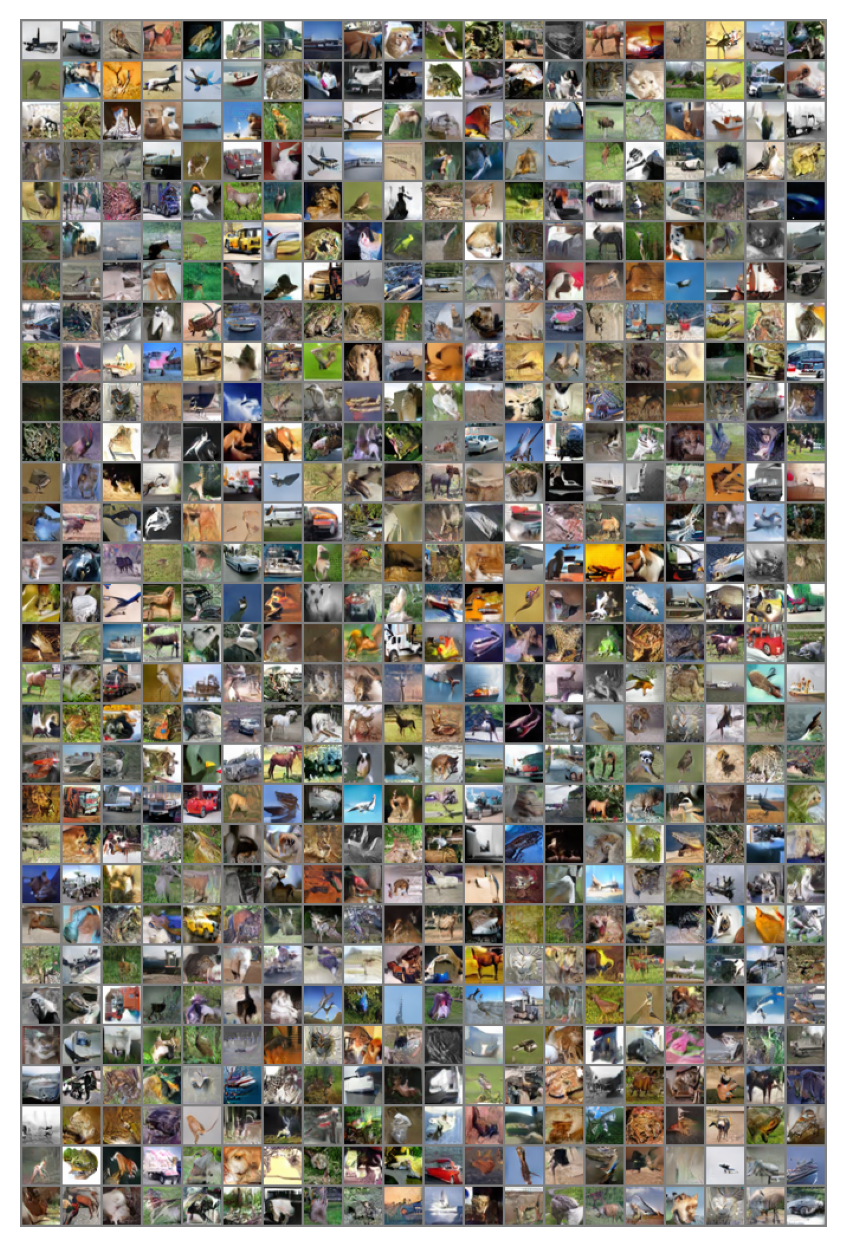}
    \caption{More CIFAR10 sample from JKO-Flow}
    \label{fig:app-cifar10}
\end{figure}

\begin{figure}
    \centering
    \includegraphics{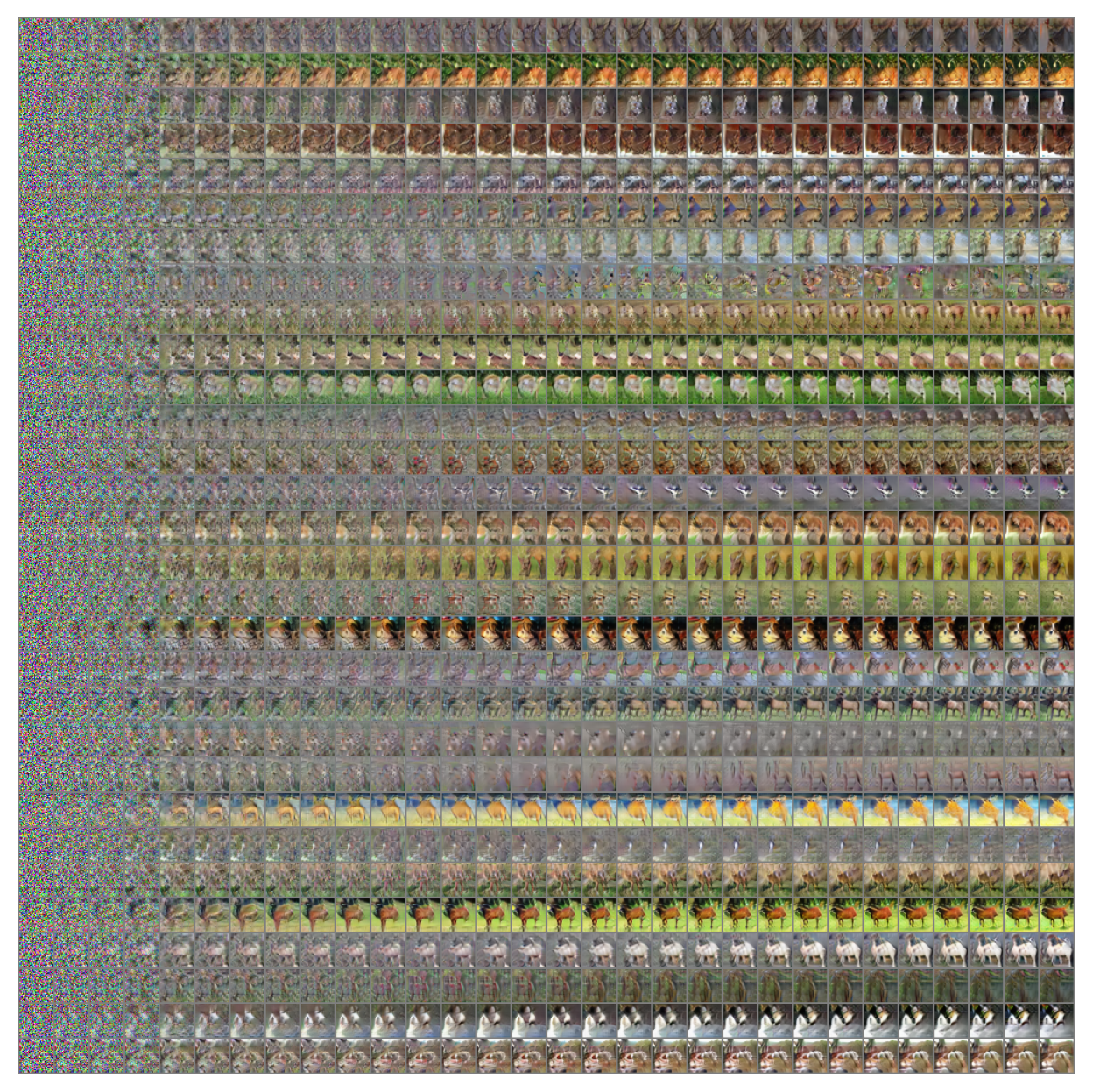}
    \caption{More CIFAR10 trajectories from JKO-Flow with $K=1$ to $K=30$.}
    \label{fig:app-cifar10-traj}
\end{figure}

We include more results of JKO-Flow. Figure~\ref{fig:app-mnist}, Figure~\ref{fig:app-cifar10}, Figure~\ref{fig:app-mnist-traj}, and Figure ~\ref{fig:app-cifar10-traj} show more generated samples from ${P_K}$ and trajectories from JKO-Flow.

\end{document}